\documentclass[a4paper,fleqn]{cas-dc}

\usepackage[authoryear]{natbib}
\usepackage{longtable}
\usepackage{booktabs}
\usepackage{array}
\usepackage{amsmath}
\usepackage{amssymb}
\usepackage{afterpage}
\usepackage{tabularx}
\usepackage{caption}
\usepackage{hyphenat}
\usepackage{algorithm}%
\usepackage{algorithmicx}%
\usepackage{algpseudocode}%
\usepackage{listings}%

\def\tsc#1{\csdef{#1}{\textsc{\lowercase{#1}}\xspace}}
\tsc{WGM}
\tsc{QE}
\begin{document}
\let\WriteBookmarks\relax
\def\floatpagepagefraction{1}
\def\textpagefraction{.001}

% Short title
\shorttitle{Efficient Search of IACs for Medical Image Segmentation}    

% Short author
\shortauthors{}  

% Main title of the paper
\title [mode = title]{Efficient Search of Implantable Adaptive Cells for Medical Image Segmentation}  

% Title footnote mark
% eg: \tnotemark[1]
% \tnotemark[1] 

% Title footnote 1.
% eg: \tnotetext[1]{Title footnote text}
% \tnotetext[1]{} 

% First author
%
% Options: Use if required
% eg: \author[1,3]{Author Name}[type=editor,
%       style=chinese,
%       auid=000,
%       bioid=1,
%       prefix=Sir,
%       orcid=0000-0000-0000-0000,
%       facebook=<facebook id>,
%       twitter=<twitter id>,
%       linkedin=<linkedin id>,
%       gplus=<gplus id>]

\author[1]{Emil Benedykciuk}[orcid=0000-0002-1542-6747]
% Footnote of the first author
% \fnmark[1]
% Corresponding author indication
\cormark[1]
% Email id of the first author
\ead{emil.benedykciuk@mail.umcs.pl}
% URL of the first author
% \ead[url]{}
% Credit authorship
% eg: \credit{Conceptualization of this study, Methodology, Software}
% \credit{}

\author[1]{Marcin Denkowski}[orcid=0000-0002-2491-091X]
% Footnote of the second author
% \fnmark[a]
% Corresponding author indication
\cormark[1]
% Email id of the second author
\ead{marcin.denkowski@mail.umcs.pl}
% URL of the second author
% \ead[url]{}
% Credit authorship
% \credit{}

\author[1]{Grzegorz M. W\'{o}jcik}[orcid=0000-0002-4678-9874]
% Footnote of the second author
% \fnmark[3]
% Email id of the second author
% \ead{grzegorz.wojcik@mail.umcs.pl}
% URL of the second author
% \ead[url]{}
% Credit authorship
% \credit{}

% Address/affiliation
\affiliation[1]{
	organization={Institute of Computer Science and Mathematics, Maria Curie Sklodowska University},
    addressline={Akademicka 9}, 
	city={Lublin},
	% citysep={}, % Uncomment if no comma needed between city and postcode
	postcode={20-033}, 
	state={},
	country={Poland}
}

% Corresponding author text
\cortext[1]{Corresponding author}

% Footnote text
% \fntext[1]{}

% For a title note without a number/mark
%\nonumnote{}

% Here goes the abstract
\begin{abstract}
	\textbf{Purpose:} Adaptive skip modules can improve medical image segmentation, but searching for them remains computationally expensive.
	Implantable Adaptive Cells (IACs) are small neural architecture search (NAS) modules inserted into U-Net skip connections, reducing the search space relative to full-network NAS for medical segmentation, yet the original IAC framework still requires a 200-epoch differentiable search for each backbone and dataset.
	
	\textbf{Methods:} We analyze the temporal behavior of operations and edges inside IAC cells during differentiable search on public medical image segmentation benchmarks.
	We observe that operations selected in the final discrete cell usually appear among the strongest candidates early in training and that the corresponding architecture parameters stabilize well before the final epoch.
	Based on this behavior, we introduce a Jensen--Shannon-divergence-based stability criterion that monitors per-edge operation-importance distributions and progressively prunes low-importance operations during search.
	The resulting accelerated framework is termed IAC-LTH.
	
	\textbf{Results:} On four public medical image segmentation benchmarks (ACDC, BraTS, KiTS, AMOS), multiple 2-D U-Net backbones, and a 2-D nnU-Net pipeline, IAC-LTH finds IAC cells whose patient-level segmentation performance matches or, in some settings, slightly exceeds that of cells obtained by the original full-length search, while reducing the NAS search budget in wall-clock time by factors between $3.7\times$ and $16\times$ across datasets and backbones.
	The effect is consistent across architectures, public benchmarks, and both non-augmented and augmented training settings, while preserving the gains of IAC-equipped U-Nets over strong attention-based and dense skip baselines.
	
	\textbf{Conclusion:} Competitive IAC architectures can be identified from early-stabilizing operations without running the full search, making adaptive skip-module design more practical for medical image segmentation under realistic computational constraints.
\end{abstract}

% Use if graphical abstract is present
%\begin{graphicalabstract}
%\includegraphics{}
%\end{graphicalabstract}

% Research highlights
\begin{highlights}
	\item Adaptive skip-module search for medical image segmentation is accelerated.
	\item IAC search cost is reduced using JS-based early operation pruning.
	\item Patient-level segmentation accuracy is preserved across four public medical datasets.
	\item The method remains effective across multiple 2-D U-Net backbones and 2-D nnU-Net.
	\item Wall-clock search time is reduced for practical model adaptation.
\end{highlights}
%\nocite{*}

% Keywords
% Each keyword is seperated by \sep

\maketitle

\section{Introduction}
\label{sec1}

Deep convolutional neural networks are now the standard tool for medical image segmentation, supporting applications such as cardiac function assessment, brain tumor delineation and multi-organ abdominal analysis \cite{bib:Survery_Taxonomy, bib:nnUNet2024, bib:cnn_for_med_survey}. 
For biomedical image processing workflows, high accuracy alone is not sufficient: segmentation models must also be stable, computationally manageable and adaptable to new datasets, scanners and protocols.
Training or redesigning large architectures from scratch is expensive and often impractical in resource-constrained clinical and research environments.

Neural Architecture Search (NAS) offers a way to automate architecture design, but in medical segmentation its practical adoption is often limited by search cost.
Differentiable NAS (DNAS) is attractive because it relaxes discrete architecture choices into continuous parameters trained by gradient descent, reducing cost relative to black-box search \cite{bib:DARTS, bib:PC-DARTS}. 
DNAS has therefore been adapted to segmentation, including medical-image-specific variants \cite{bib:IAC, bib:NAS-Unet, bib:DiNTS, bib:HyperSegNAS, bib:DIPO, bib:DAST, bib:HASA}.

In previous work, Implantable Adaptive Cells (IACs) were proposed as a NAS-based plug-in module for pre-trained U-Nets \cite{bib:IAC}.
Instead of searching the entire encoder-decoder topology, IAC searches only over small directed-acyclic-graph (DAG) cells inserted into the skip connections.
The backbone U-Net is trained first and then frozen; NAS is applied to the skip modules, and finally the selected cells are fine-tuned while the backbone remains fixed \cite{bib:IAC}.
This ``retrofit'' strategy significantly reduces search cost compared with full-network NAS and yields consistent Dice gains over plain U-Nets, attention-based skips and dense skip variants across several medical benchmarks.

Despite this efficiency, the original IAC framework still runs a full differentiable search for 200 epochs per dataset and backbone.
Early ablations hinted that competitive cells can appear much earlier, but the search protocol remained conservative and did not exploit these observations.

To address this practical bottleneck, we examine whether the search can be shortened once the operation choices inside an IAC cell have effectively stabilized.
This viewpoint is related to the Lottery Ticket Hypothesis (LTH), which suggests that strong subnetworks can emerge early during optimization \cite{bib:LTH}.
In our setting, the key question is not sparse training in general, but whether early-stabilizing operations inside skip modules can be identified reliably enough to reduce search cost without weakening medical segmentation performance.

\begin{figure*}[ht!]
	\centering
	\includegraphics[width=0.8\textwidth]{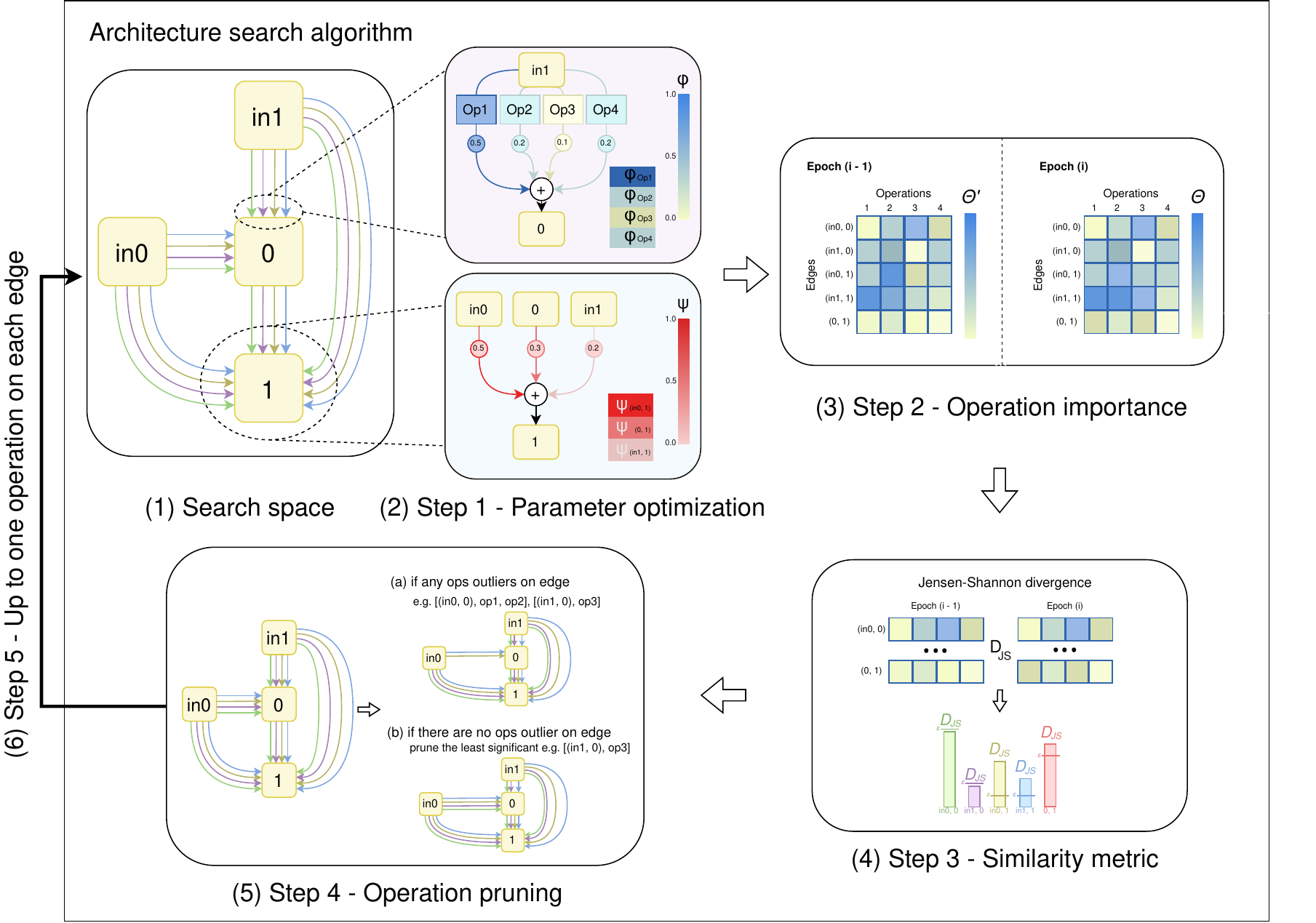}
	\caption{Overview of the proposed IAC search algorithm.
		(1) Search-space definition: each iteration starts from the current set of candidate operations on every edge. 
		(2) Joint optimization: we update both the architecture parameters and the cell weights.
		(3) Operation influence: after every epoch we compute the influence score of each operation. 
		(4) Stability test: we measure the Jensen--Shannon divergence between the influence vectors from the current and the previous epoch; if the change on an edge is below its edge-specific threshold, we move to pruning. 
		(5) Edge-wise pruning: (5.a) if one or more operations are at least two standard deviations below the mean influence on that edge, we remove those outliers; (5.b) otherwise, we remove the single operation with the lowest influence score on that edge. 
		(6) Convergence: we repeat steps 2-5 until each edge keeps exactly one operation, then select the two most influential incoming edges for every node.
		See Section~\ref{sec:method} and Algorithm~\ref{algs:alg1} for full details.}
	\label{fig:schema}
\end{figure*}

This paper investigates operation and edge stabilization inside IAC cells and asks whether early ``winning'' operations can be used to shorten the search while preserving segmentation performance.
To answer this, we first perform a detailed analysis of the temporal dynamics of IAC search on four public medical segmentation benchmarks (ACDC, BraTS, KiTS, AMOS) and multiple 2-D U-Net backbones.
We track operation and edge weights over time, quantify how fast the architecture stabilizes, and measure when the final winning operations first appear among the top candidates.
Building on these insights, we propose a Jensen--Shannon (JS) divergence-based stability criterion that allows us to prune low-importance operations and edges on-the-fly, focusing training on early winners and reducing search time by factors of up to an order of magnitude, without materially changing Dice.
Figure~\ref{fig:schema} sketches the workflow of the proposed method.

Our main contributions are as follows:
\begin{itemize}
	\item \textbf{An efficient search strategy for adaptive skip modules in medical image segmentation.}
	We show across four public datasets and several 2-D U-Net backbones that IAC search can be shortened while preserving the segmentation gains of implantable skip modules.
	\item \textbf{Operation-level analysis of IAC search dynamics.}
	By representing genotypes as operation-edge importance vectors and comparing them over epochs, we show that most final winning operations appear among the top candidates very early and that the continuous architecture parameters converge rapidly.
	\item \textbf{A JS-divergence-based stability test for differentiable search.}
	We monitor the Jensen--Shannon divergence between operation-importance distributions on each edge across epochs.
	Small divergence indicates that an edge has stabilized, providing a principled criterion for when it is safe to prune low-importance operations and avoid further exploration.
	\item \textbf{An efficient search algorithm for IACs (IAC-LTH).}
	Building on the stability test, we define a progressive pruning scheme that removes operations falling below a dataset-adaptive importance threshold on stable edges, and ultimately prunes entire edges once they repeatedly fail to reach non-zero importance.
	Search stops when each node retains exactly two incoming edges with a single operation on each, after which the resulting cell is discretized and trained as in the original IAC pipeline.
	This reduces the effective search budget by factors between approximately $3.7\times$ and $16\times$, while preserving the underlying optimization framework.
	\item \textbf{Consistent patient-level segmentation performance at reduced search cost.}
	Across ACDC, BraTS, KiTS and AMOS, the accelerated search yields IAC-equipped U-Nets whose Dice scores are on par with or slightly better than those obtained with the full 200-epoch IAC search, while the wall-clock search time is reduced.
	The method also remains competitive with strong attention-gated and dense skip baselines under identical training protocols and extends to a 2-D nnU-Net pipeline, including augmented training.
	\item \textbf{A resource-aware retrofit strategy for medical segmentation.}
	Our analysis and algorithm show that reliable IAC cells can be discovered by focusing on early winning operations and monitoring JS-based stability, making adaptive skip-module search feasible in settings where full-length searches would otherwise be difficult to justify.
	Our implementation is publicly available at:\\
	\url{https://gitlab.com/emil-benedykciuk/u-net-darts-tensorflow/-/tree/lth-analysis}.
\end{itemize}

\section{Related Work}

\subsection{NAS for Medical Image Segmentation and Skip Connections}

Here we focus on works that are most relevant to our setting: medical image segmentation methods that search or redesign U-Net-like encoder-decoder architectures, especially through skip-connection modules.

Several methods search full encoder-decoder topologies or large building blocks. 
Auto-DeepLab \cite{bib:AutoDeepLab} and FasterSeg \cite{bib:FasterSeg} optimize both backbone and decoder for semantic segmentation. 
For medical images, NAS-Unet \cite{bib:NAS-Unet}, DiNTS \cite{bib:DiNTS}, HyperSegNAS \cite{bib:HyperSegNAS} and related approaches \cite{bib:DIPO, bib:DAST, bib:HASA} apply DNAS or other NAS methods to find encoder-decoder designs and multi-scale fusion strategies. 
These methods can produce strong models, but the search spaces are relatively large and the associated computational cost can limit practical use in biomedical segmentation studies.

An alternative line of work restricts NAS to smaller plug-in modules.
Instead of redesigning the entire network, the search focuses on blocks that can be inserted into a strong baseline. 
Implantable Adaptive Cells follow this strategy: small differentiable cells are searched only in U-Net skip connections, while the backbone is pre-trained and then frozen \cite{bib:IAC}. 
This retrofit paradigm significantly reduces search cost and yields consistent improvements over plain, attention-based and dense skip connections on several medical datasets.
The importance of carefully designed skip connections is further highlighted in the survey of Xu et al., who analyze their role and evolution in modern segmentation networks \cite{bib:survey_skip}.

The present work builds directly on the IAC framework and addresses its main practical limitation: the original method still runs a full 200-epoch differentiable search per backbone and dataset. 
We analyze how quickly the IAC architecture stabilizes during search and propose a pruning scheme, termed IAC-LTH, that shortens the search by factors between approximately $3.7\times$ and $16\times$ while preserving the segmentation gains of IACs on public medical benchmarks.

\subsection{Early Stabilization, Pruning, and NAS}

The Lottery Ticket Hypothesis \cite{bib:LTH,bib:LTH_survey} has motivated a large body of work on discovering sparse subnetworks that match the performance of dense models. 
Classical pipelines rely on iterative magnitude pruning and repeated train-prune-retrain cycles, which are effective but computationally demanding. 
One-shot pruning methods such as SNIP and GraSP compute saliency or gradient-flow scores at or near initialization and remove low-score weights in a single step \cite{bib:SNIP,bib:GraSP}. 
Although attractive in terms of cost, these methods typically underperform data-driven pruning on difficult tasks.

To reduce cost further, several works focus on early ticket detection. 
Early-Bird Tickets \cite{bib:early_birds} show that the pruning mask often stabilizes after a fraction of training epochs. 
By monitoring the Hamming distance between successive masks and stopping dense training once the distance vanishes, the method saves up to several times the compute without loss in accuracy. 
Other studies report similar mask-stabilization phenomena in object detection and other domains \cite{bib:obj_reco_lth}. 
These results suggest that the topology of a winning subnetwork emerges early and that later epochs mostly refine an already identified structure.

NAS and LTH have also been combined more tightly. 
SuperTickets \cite{bib:supertickets} integrate architecture search and pruning within a single supernet: connections and channels are ranked and pruned during search, producing compact subnetworks that match or surpass traditional NAS followed by post-hoc pruning. 
At the operation level, several works propose alternatives to naive magnitude-based ranking in DNAS. 
For example, Wang et al.\ measure the drop in validation accuracy when an operation is disabled \cite{bib:rethinking_dnas}, and Xiang et al.\ employ zero-cost metrics to score individual operations, accelerating DARTS by up to tens of times \cite{bib:zero_cost_op_score_dnas}. 
These methods treat operation selection itself as a structured lottery, where some operations can be removed early without harming final performance.

Structured pruning generalizes LTH from unstructured weights to filters, channels and operations. 
Alabdulmohsin et al.\ proposed a generalized LTH that covers both unstructured and structured sparsity \cite{bib:generalized_lth}, supporting the view that entire operations or paths may form the essential core of a model. 
Our work adopts this perspective in a medical segmentation setting: we investigate operations and edges inside an IAC cell and show that the operations that survive discretization typically emerge among the top candidates after only a small fraction of the search budget.

\subsection{Stability Metrics and Divergence-Based Early Stopping}

A central question in applying LTH ideas during NAS is how to decide when the architecture has effectively stabilized. 
Early-Bird Tickets rely on \emph{mask distance}, the Hamming distance between binary pruning masks at successive checkpoints, and stop dense training once this distance becomes zero \cite{bib:early_birds}. 
Kim et al.\ introduced Early-Time, which monitors the Kullback--Leibler (KL) divergence between output distributions and terminates training when the divergence falls below a threshold \cite{bib:early_time_lth}. 
Both metrics serve as proxies for the emergence of a stable subnetwork.

Similar stability concepts appear in differentiable NAS. 
Shen et al.\ proposed the Early Pruning Indicator (EPI), which tracks changes in the architecture parameters of DARTS-like supernets and prunes operations once their selection probabilities stop moving significantly \cite{bib:when_to_prune}. 
Zero-shot and training-free NAS can be viewed as extreme cases where stability is assumed at initialization and candidate operations are ranked using proxy scores \cite{bib:zero_shot_NAS,bib:zero_cost_op_score_dnas}. 

Information-theoretic measures, such as KL and JS divergence, are natural tools for quantifying stability of probability distributions over operations or masks. 
KL divergence is asymmetric and unbounded, which complicates threshold selection, whereas JS divergence is symmetric and bounded. 
In this work we use per-edge JS divergence between successive operation-importance distributions inside the IAC cell to detect when an edge has stabilized. 
Once JS divergence on an edge remains below a threshold for several checks, we prune low-importance operations and, eventually, entire edges. 
This implements an operation-level analogue of Early-Bird and EPI criteria directly inside the IAC search, tailored to the skip-only medical segmentation setting.

\subsection{Positioning of This Work}

In summary, prior work has shown that NAS can design strong segmentation architectures, that pruning-inspired ideas can reduce search cost, and that stability metrics enable early detection of reliable subnetworks. 
However, to the best of our knowledge, these ideas have not been systematically explored inside implantable cells and other retrofit NAS methods that modify skip connections of pre-trained U-Nets for medical image segmentation.

Our contribution is to fill this gap from the perspective of practical biomedical image segmentation. 
We analyze operation-level stabilization inside IAC cells, quantify how quickly their architecture parameters stabilize across four medical datasets and multiple U-Net backbones, and introduce a JS-divergence-based stability test that drives on-the-fly pruning of operations and edges. 
This leads to an accelerated IAC search (IAC-LTH) that preserves or improves the segmentation gains of IACs while reducing search time in wall-clock terms by factors between approximately $3.7\times$ and $16\times$.

Compared to the original IAC framework \cite{bib:IAC}, this paper makes three main additions:
(i) an operation-level analysis of IAC search dynamics, quantifying how quickly winning operations stabilize;
(ii) a Jensen--Shannon-divergence stability criterion with a concrete pruning rule that progressively removes low-importance operations and edges; and 
(iii) an extensive evaluation of the accelerated search both in a frozen U-Net backbone grid and inside the nnU-Net pipeline, showing preserved or improved segmentation performance at $3$-$16\times$ lower search cost.

\section{Baseline: Implantable Adaptive Cells and Skip-Only NAS}
\label{sec:baseline_iac}

In this section we briefly summarize the original Implantable Adaptive Cell design and search procedure, which form the basis for our analysis and the proposed accelerated search.

\subsection{IAC topology and placement}

The IAC is a small module based on a directed acyclic graph $G = (V, E)$ that is inserted into the skip connections of a U-Net to improve feature fusion between the encoder and decoder \cite{bib:IAC}. 
The node set $V$ contains $N$ nodes, indexed from $0$ to $N-1$, and the edge set $E$ contains all directed candidate edges $(i,j)$ with $i < j$.

The first two nodes $v_0$ and $v_1$ are input nodes:
$v_0$ receives encoder features (skip tensor) and $v_1$ receives decoder features (upsampled tensor) at a given resolution level. 
Both inputs are first passed through $1\times1$ convolutions to match the internal channel dimension of the cell so that the same IAC genotype can be plugged into any U-Net configuration.

The remaining nodes $v_2,\dots,v_{N-1}$ are intermediate computation nodes. 
The output of the cell is obtained by concatenating the outputs of the last $R$ nodes,
\begin{equation}
	x_{\text{out}} = \mathrm{Concat}(x_{N-R}, x_{N-R+1}, \dots, x_{N-1}),
\end{equation}
followed by a $1\times1$ convolution that maps the concatenated features to the number of channels expected by the U-Net decoder at that scale.

At each scale of the U-Net, the original identity skip connection between encoder and decoder is replaced by such an IAC. 
The same discrete cell topology (genotype) is reused across all skip levels, but each instance has its own weights.

\subsection{Edge transforms and PC-DARTS-style mixed operations}
\label{sec:baseline_edge_transform}

Each edge $(i,j) \in E$ with $i < j$ carries a mixed operation that transforms the feature tensor $x_i$ at node $v_i$ into a contribution to node $v_j$. 
The mixed operation is defined over a fixed set of candidate operations $O$:
identity, $3\times3$ max pooling, $3\times3$ average pooling, $3\times3$/$5\times5$ separable convolution, and $3\times3$/$5\times5$ dilated separable convolution.

For each edge $(i,j)$ and operation $o \in O$ we maintain an architecture parameter $\alpha^o_{i,j}$. 
During search, these parameters are transformed into operation probabilities using the softmax function:
\begin{equation}
	\phi^o_{i,j} = \frac{\exp(\alpha^o_{i,j})}{\sum_{o' \in O} \exp(\alpha^{o'}_{i,j})}.
	\label{eq:phi_softmax}
\end{equation}
The basic mixed operation on all channels would be
\begin{equation}
	f_{i,j}(x_i) = \sum_{o \in O} \phi^o_{i,j} \, o(x_i).
\end{equation}

To reduce memory and computation cost, the IAC search adopts the partial-channel connection mechanism from PC-DARTS \cite{bib:PC-DARTS}. 
Let $M_{i,j}$ be a binary channel mask that selects a fraction $1/K$ of the channels in $x_i$. 
The masked channels are processed by the mixed operation, while the remaining channels are copied through unchanged:
\begin{equation}
	f^{\mathrm{PC}}_{i,j}(x_i; M_{i,j}) = 
	\sum_{o \in O} \big(\phi^o_{i,j} \, o(M_{i,j} \odot x_i)\big) 
	+ (1 - M_{i,j}) \odot x_i,
	\label{eq:pc_op}
\end{equation}
where $\odot$ denotes element-wise multiplication with broadcasting over spatial dimensions. 
Thus, only a subset of channels is transformed by candidate operations, while the rest provide a cheap bypass.
Following \cite{bib:PC-DARTS, bib:IAC}, we set $K = 4$.

To stabilize the competition between incoming edges at each node, a second set of architecture parameters $\beta_{i,j}$ is introduced to control edge importance. 
For node $j$ the raw edge logits $\beta_{i,j}$ for all $i < j$ are first transformed using a monotonic non-linearity and then normalized by softmax:
\begin{equation}
	\hat{\Psi}_{i,j} = 
	\frac{\exp\big(\tan(\beta_{i,j}) - \max_{k<j}\tan(\beta_{k,j})\big)}
	{\sum_{k<j} \exp\big(\tan(\beta_{k,j}) - \max_{k'<j}\tan(\beta_{k',j})\big)}.
	\label{eq:psi_hat}
\end{equation}
The subtraction of the maximum ensures numerical stability.

To prevent any single edge from dominating the optimization and to encourage sparse connectivity, \cite{bib:IAC} further apply a capping scheme. 
If the largest normalized weight satisfies $\max_{i<j} \hat{\Psi}_{i,j} \leq 0.5$, we simply set
\begin{equation}
	\Psi_{i,j} = \hat{\Psi}_{i,j}, \quad \text{if } \max_{i<j} \hat{\Psi}_{i,j} \leq 0.5.
	\label{eq:psi_uncapped}
\end{equation}
Otherwise, let $i^\ast = \arg\max_{i<j} \hat{\Psi}_{i,j}$ and denote the sum of the remaining weights by
\begin{equation}
	S_{\text{rest}} = \sum_{\substack{k<j\\k \neq i^\ast}} \hat{\Psi}_{k,j}.
\end{equation}
The capped edge weights are then defined as
\begin{equation}
	\Psi_{i,j} =
	\begin{cases}
		0.5, & \text{if } i = i^\ast,\\[2mm]
		0.5 \, \dfrac{\hat{\Psi}_{i,j}}{S_{\text{rest}}}, & \text{if } i \neq i^\ast.
	\end{cases}
	\label{eq:psi_capped}
\end{equation}
This projection ensures that at most one edge per node can carry more than half of the total incoming weight, discouraging degenerate single-edge solutions while preserving differentiability.

Given the edge weights $\Psi_{i,j}$, the output of node $j$ is a weighted sum of all incoming partial-channel mixed operations:
\begin{equation}
	x^{\mathrm{PC}}_j = \sum_{i<j} \Psi_{i,j} \, f^{\mathrm{PC}}_{i,j}(x_i; M_{i,j}).
	\label{eq:node_update}
\end{equation}

\subsection{Continuous relaxation and bilevel optimization}
\label{sec:baseline_bilevel}

To enable differentiable optimization of the architectural parameters $(\alpha, \beta)$ together with the cell weights $\omega$, the IAC search adopts the standard continuous-relaxation paradigm. 
During search, all edges $(i,j)$ are active and contribute according to \eqref{eq:pc_op}-\eqref{eq:node_update}. 
Let $\omega_U$ denote the weights of the frozen base U-Net and $\omega$ the weights of the IAC cell.

The search objective can be written as a bilevel optimization problem:
\begin{gather}
	\min_{\alpha,\beta} \; \mathcal{L}_{\mathrm{val}}\big(\omega(\alpha,\beta,\omega_U), \alpha, \beta, \omega_U\big),
	\label{eq:bilevel_outer}\\
	\text{s.t. } \quad 
	\omega(\alpha,\beta,\omega_U) = 
	\arg\min_{\omega} \; \mathcal{L}_{\mathrm{train}}(\omega, \alpha, \beta, \omega_U),
	\label{eq:bilevel_inner}
\end{gather}
where $\mathcal{L}_{\mathrm{train}}$ and $\mathcal{L}_{\mathrm{val}}$ are the training and validation losses, respectively.

In practice, following \cite{bib:IAC}, a first-order alternating approximation is used for efficiency. 
Search is run for a fixed number of epochs.
Within each epoch, we alternate between:
(i) updating the cell weights $\omega$ by minimizing $\mathcal{L}_{\mathrm{train}}$ on the training split, and 
(ii) updating the architecture parameters $(\alpha,\beta)$ by minimizing $\mathcal{L}_{\mathrm{val}}$ on the validation split. 
The U-Net backbone weights $\omega_U$ remain frozen throughout the search.

\subsection{Discretization and final training}
\label{sec:baseline_discretization}

After the continuous search process concludes, the architecture parameters $(\alpha,\beta)$ are discretized to obtain the final IAC genotype.

On each edge $(i,j)$, the operation with the highest probability $\phi^o_{i,j}$ in \eqref{eq:phi_softmax} is selected:
\begin{equation}
	o_{i,j} = \arg\max_{o \in O} \phi^o_{i,j}.
\end{equation}

Next, for each node $j$ we retain exactly two incoming edges with the largest edge weights $\Psi_{i,j}$ as defined in \eqref{eq:psi_uncapped}-\eqref{eq:psi_capped}, in order to limit the complexity of the discrete cell. 
The resulting sparse DAG defines the discrete IAC architecture.

This discrete cell is then instantiated in the U-Net that was used during search: all skip connections are replaced by the discrete IAC, and the cell weights $\omega$ are trained (fine-tuned) while keeping the backbone weights $\omega_U$ frozen.
The training loss and optimizer follow the same configuration as in the baseline U-Net.
This three-stage pipeline: (i) backbone pre-training, (ii) IAC search with frozen backbone, and (iii) final training of the discrete IAC, constitutes the baseline on top of which we perform our Lottery-Ticket-based analysis and define the accelerated search method IAC-LTH.

\section{Analysis}
\label{sec:analysis}
In this section we analyze the evolution of the original IAC architecture during differentiable search. 
Building on the baseline search procedure described in Section~\ref{sec:baseline_iac}, we treat the architecture parameters $(\alpha,\beta)$ on each edge as a proxy for the importance of candidate operations and ask:

\begin{enumerate}
	\item How quickly do the operations and edges that survive in the final discrete IAC emerge during search?
	\item How different are the architectures immediately after warm-up, in the middle of the search, and at the end?
\end{enumerate}

We first summarize the experimental setting used for this analysis and then introduce the genotype representations and similarity measures that underpin the figures in this section.

\subsection{Experimental setting for the analysis}
\label{sec:analysis_setup}

The analysis is performed on four public medical segmentation datasets: ACDC~\cite{bib:acdc}, BraTS~\cite{bib:BraTS-2015,bib:BraTS-Dodatkowy1,bib:BraTS-2023}, KiTS~\cite{bib:kits21} and AMOS~\cite{bib:amos}. 
For each dataset we follow a patient-wise split into training (60\%), validation (20\%) and test (20\%) sets. 
The training set is further divided once into \textit{train\_search} and \textit{val\_search} of equal size for the IAC search, following the original method~\cite{bib:IAC}; this split is fixed for all experiments.
The  \textit{train\_search}/ \textit{val\_search} split described above is used for analyzing the original IAC search in this section; in the proposed IAC-LTH algorithm we instead optimize on the full training set without an internal search/validation split (Section~\ref{sec:method}).

Unless otherwise stated, all experiments use a 2-D U-Net~\cite{bib:unet} as backbone with one of several encoders (VGG-16~\cite{bib:vgg16}, ResNet-50~\cite{bib:ResNet}, MobileNetV3-Large~\cite{bib:mobilenetV3}, EfficientNetV2-S/M~\cite{bib:effv2}). 
Input slices and segmentation masks are cropped to $128\times128$ pixels; each input channel corresponds to an imaging modality (e.g.\ T1, T1Gd, T2, FLAIR for BraTS) and each output channel to a foreground class.
Pre-processing (normalization, standardization) is identical for the baseline U-Nets and IAC-equipped models and follows the protocol described in Section~\ref{sec:experiments_setup}.

To isolate architectural effects and reduce randomness, we deliberately \emph{disable data augmentation} during both IAC search and model training in this analysis. 
This choice standardizes the comparison between reference U-Nets and IAC-enhanced models but may also make the baselines more prone to overfitting on small datasets. 
If IAC-equipped models still achieve superior test Dice under these conditions, this provides strong evidence that the gains come from better architecture rather than from stochastic regularization.
In the full nnU-Net setting with augmentation, the IAC and accelerated-search effects are evaluated separately in Section~\ref{sec:experiments}.

During the warm-up and search phases, the IAC weights $\omega$ are optimized using SGD (learning rate $0.01$, no momentum, no weight decay), while the architecture parameters $(\alpha,\beta)$ are optimized with Adam (learning rate $0.001$). 
Random seeds are fixed for all relevant TensorFlow and CUDA operations as far as supported. 
All experiments are implemented in Python~3.9 using TensorFlow~2.17 and run on an NVIDIA RTX~3090~Ti GPU.
All configuration files and random seeds are provided in the accompanying repository.

\subsection{Genotype representations and similarity measures}
\label{sec:analysis_genotypes}

As described in Section~\ref{sec:baseline_edge_transform}, each edge $(i,j)$ in the IAC cell maintains architecture parameters $\alpha^o_{i,j}$ for every operation $o \in O$ and an edge parameter $\beta_{i,j}$. 
For analysis, we define the \emph{importance} of operation $o$ on edge $(i,j)$ as
\begin{equation}
	\Theta^o_{i,j} = \alpha^o_{i,j} \cdot \beta_{i,j},
\end{equation}
that is, the product of the operation-specific and edge-specific logits.
Collecting all $\Theta^o_{i,j}$ into a single vector yields the \emph{continuous genotype} at epoch $e$,
\begin{equation}
	\boldsymbol{\Theta}^{(e)} \in \mathbb{R}^{|\mathcal{E}||O|},
\end{equation}
where $\mathcal{E}$ is the set of candidate edges. 
For the IAC cell used here this results in a 98-dimensional vector (14 edges $\times$ 7 operations).

To study how architectures change over time, we consider several discrete views of the genotype:

\begin{itemize}
	\item \textbf{Final discrete genotype}: after search, each edge keeps the operation with highest importance and each node keeps its two strongest incoming edges (as in Section~\ref{sec:baseline_discretization}). 
	This discrete topology is encoded as a binary vector $\mathbf{g}^{\text{final}}$ with the same indexing as $\boldsymbol{\Theta}^{(e)}$.
	\item \textbf{Top-$k$ discrete masks}: at each epoch $e$, we construct binary vectors $\mathbf{g}^{(e)}_{(k)}$ for $k\in\{1,2,3\}$ where $\mathbf{g}^{(e)}_{(k)}(i,j,o)=1$ if operation $o$ is among the $k$ highest-scoring operations on edge $(i,j)$ according to $\Theta^o_{i,j}$ and $0$ otherwise.
\end{itemize}

\begin{figure*}[ht!]
	\centering
	\includegraphics[width=1\textwidth]{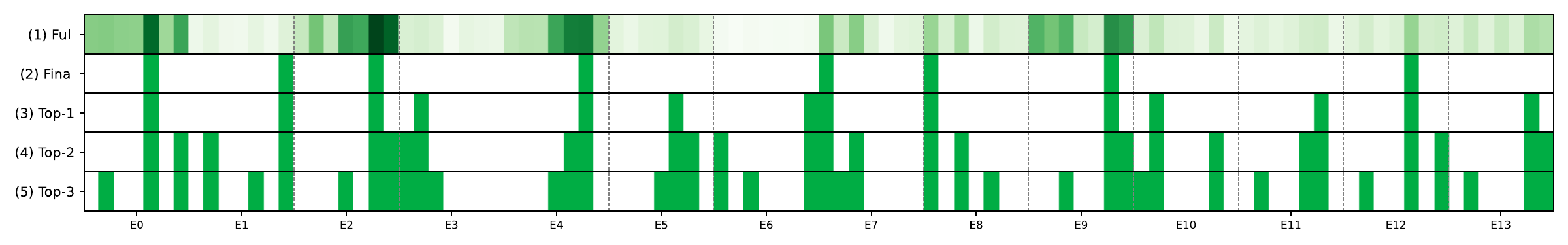}
	\caption{Example search-vector representations.
		(1) the full operation-importance vector (continuous values, $\alpha \cdot \beta$ for each operation on each edge), (2) the discrete representation of the final architecture, (3) a discrete representation where each edge is assigned its top-1 operation at that stage, (4) one with the top-2 operations per edge, and (5) one with the top-3 operations per edge.}
	\label{fig:3}
\end{figure*}

These representations are illustrated in Figure~\ref{fig:3}.
We use two similarity measures:

\textbf{Cosine similarity:} For any two epochs $e_1,e_2$ we define
\begin{equation}
	\mathrm{CosSim}(e_1,e_2) = 
	\frac{\boldsymbol{\Theta}^{(e_1)} \cdot \boldsymbol{\Theta}^{(e_2)}}{
		\big\|\boldsymbol{\Theta}^{(e_1)}\big\|_2 \,\big\|\boldsymbol{\Theta}^{(e_2)}\big\|_2}.
\end{equation}
This captures global alignment of operation importance over time.

\textbf{Hamming similarity:} For discrete vectors (final genotype or top-$k$ masks) we use Hamming similarity,
\begin{equation}
	\mathrm{HamSim}(e_1,e_2;k) 
	= 1 - \frac{1}{|\mathcal{E}||O|} \sum_{i,j,o} 
	\mathbb{1}\big[\mathbf{g}^{(e_1)}_{(k)}(i,j,o) \neq \mathbf{g}^{(e_2)}_{(k)}(i,j,o)\big],
\end{equation}
where $\mathbb{1}[\cdot]$ is the indicator function. 
This measures the fraction of entries in the binary masks that agree between epochs.

We analyze similarities across the entire search trajectory and at specific checkpoints: after warm-up (epoch 15) and at epochs 16, 20, 40, 80, 100, 150 and 200.

\subsection{Qualitative evolution of operation importance}

\begin{figure*}[ht!]
	\centering
	\includegraphics[width=0.8\textwidth]{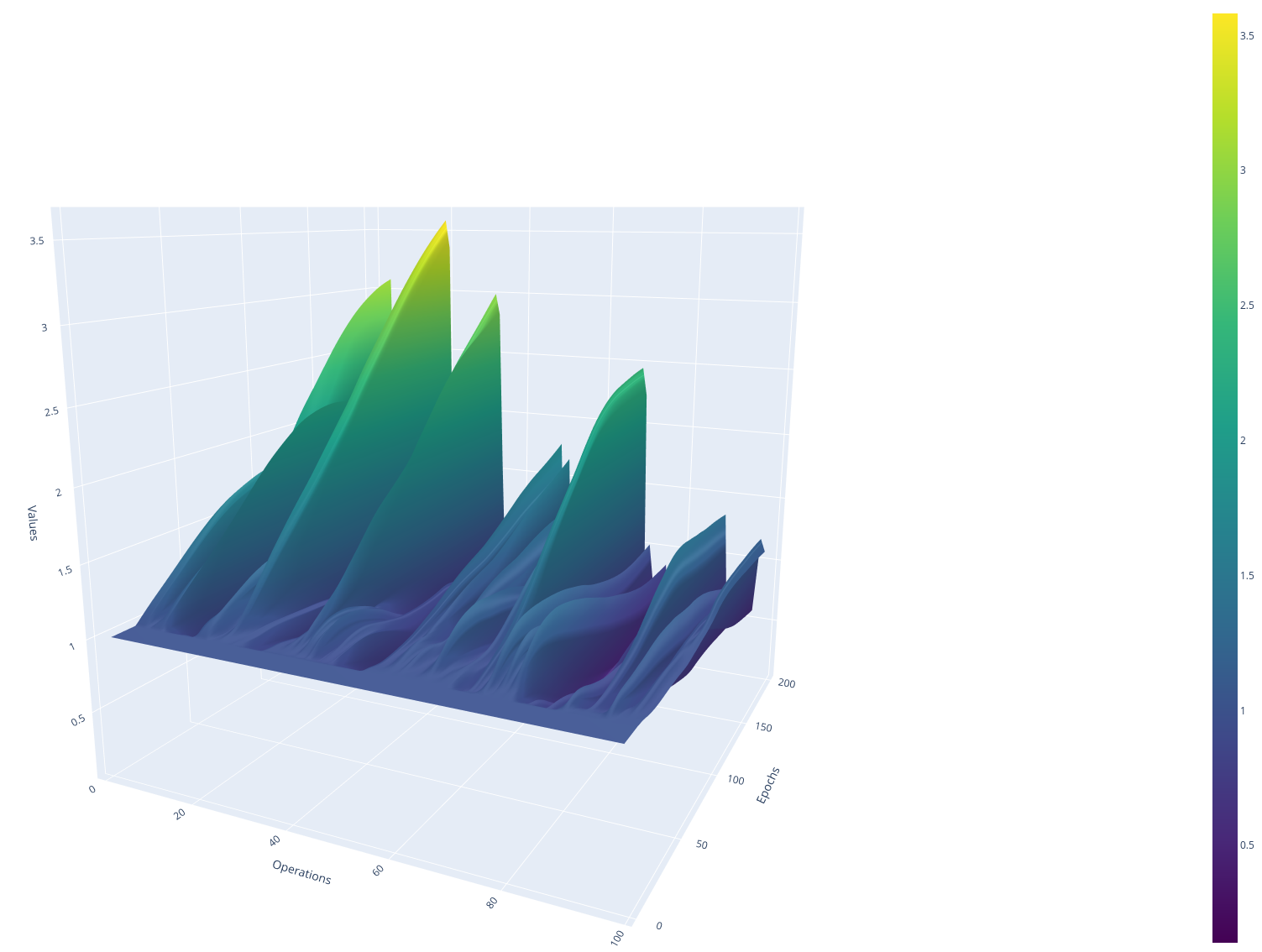}
	\caption{Evolution of the importance of operations during the search process (epochs).
		The importance of an operation on edge $(i,j)$ is defined as the product of the operation-specific logit $\alpha^o_{i,j}$ and the edge-specific logit $\beta_{i,j}$.}
	\label{fig:2}
\end{figure*}

Figure~\ref{fig:2} visualizes the evolution of operation importance over search epochs for a representative run. 
Each colored trajectory corresponds to the importance $\Theta^o_{i,j}$ of one operation on one edge.

Most final-winning operations become clearly dominant early in the search: their importance increases rapidly after the warm-up and then remains high and stable. 
In contrast, many other operations are quickly suppressed and never regain prominence. 
This confirms earlier performance-level observations from~\cite{bib:IAC}: the main architectural variability is concentrated at the beginning of the search.

\subsection{Similarity of architectures over search time}

\begin{figure*}[ht!]
	\centering
	\includegraphics[width=0.9\textwidth]{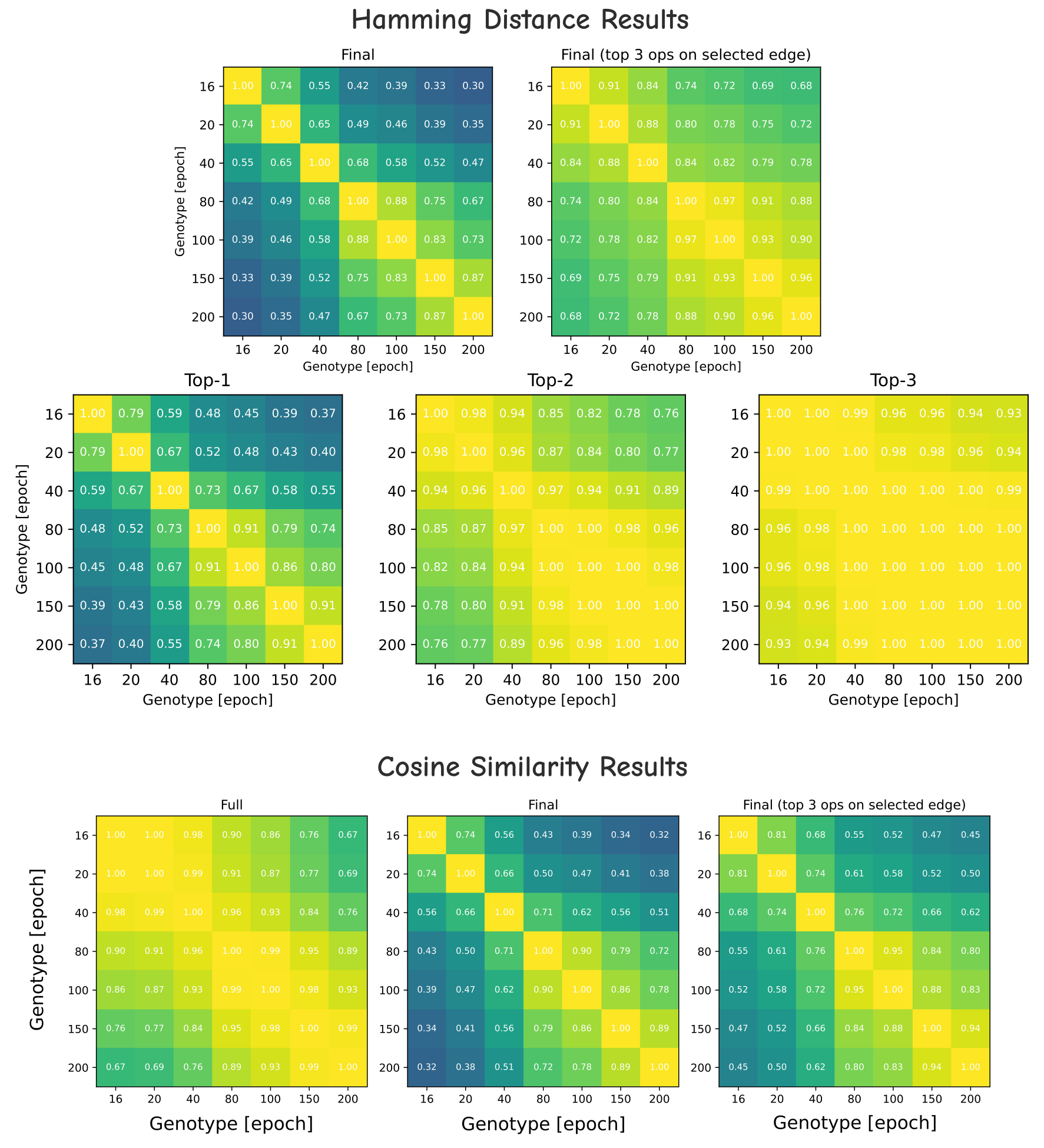}
	\caption{Similarity matrices between IAC genotypes sampled at different epochs.
		Each matrix entry compares two epochs using either Hamming similarity (discrete rows) or cosine similarity (continuous rows) of the genotype vectors introduced in Figure~\ref{fig:3}. 
		\emph{Final} -- similarity to the discretized genotype used for deployment: every node keeps its two strongest incoming edges, and each edge keeps the single operation with the highest importance. 
		\emph{Full} -- similarity to the complete continuous genotype, i.e.\ the full 98-D vector of $\alpha \cdot \beta$ values. 
		\emph{Top-1/Top-2/Top-3} -- similarity computed after retaining, on every edge, only the 1, 2 or 3 operations with the largest $\alpha \cdot \beta$ scores, as defined in Figure~\ref{fig:3}. 
		Light cells mark high similarity (stable architecture), dark cells mark low similarity (architecture still evolving).}
	\label{fig:4}
\end{figure*}

Based on the genotype representations in Figure~\ref{fig:3}, Figure~\ref{fig:4} shows similarity matrices between architectures sampled at different epochs. 
Rows correspond to different representations (full continuous vector, top-1/2/3 discrete vectors, final discrete genotype), and columns to source and target epochs.

For the continuous genotypes (Full row), cosine similarity between different epochs grows quickly: the matrices exhibit bright blocks after the first 40-60 epochs, indicating that $\boldsymbol{\Theta}^{(e)}$ changes only slightly afterwards. 
The discrete top-3 and top-2 representations stabilize somewhat later, while the top-1 and final discrete genotype show non-trivial changes almost until the end of the search. 
This is expected: small shifts in importance near decision boundaries can swap the top-1 candidate even when the underlying distribution is already stable.

\begin{figure*}[ht!]
	\centering
	\includegraphics[width=0.8\textwidth]{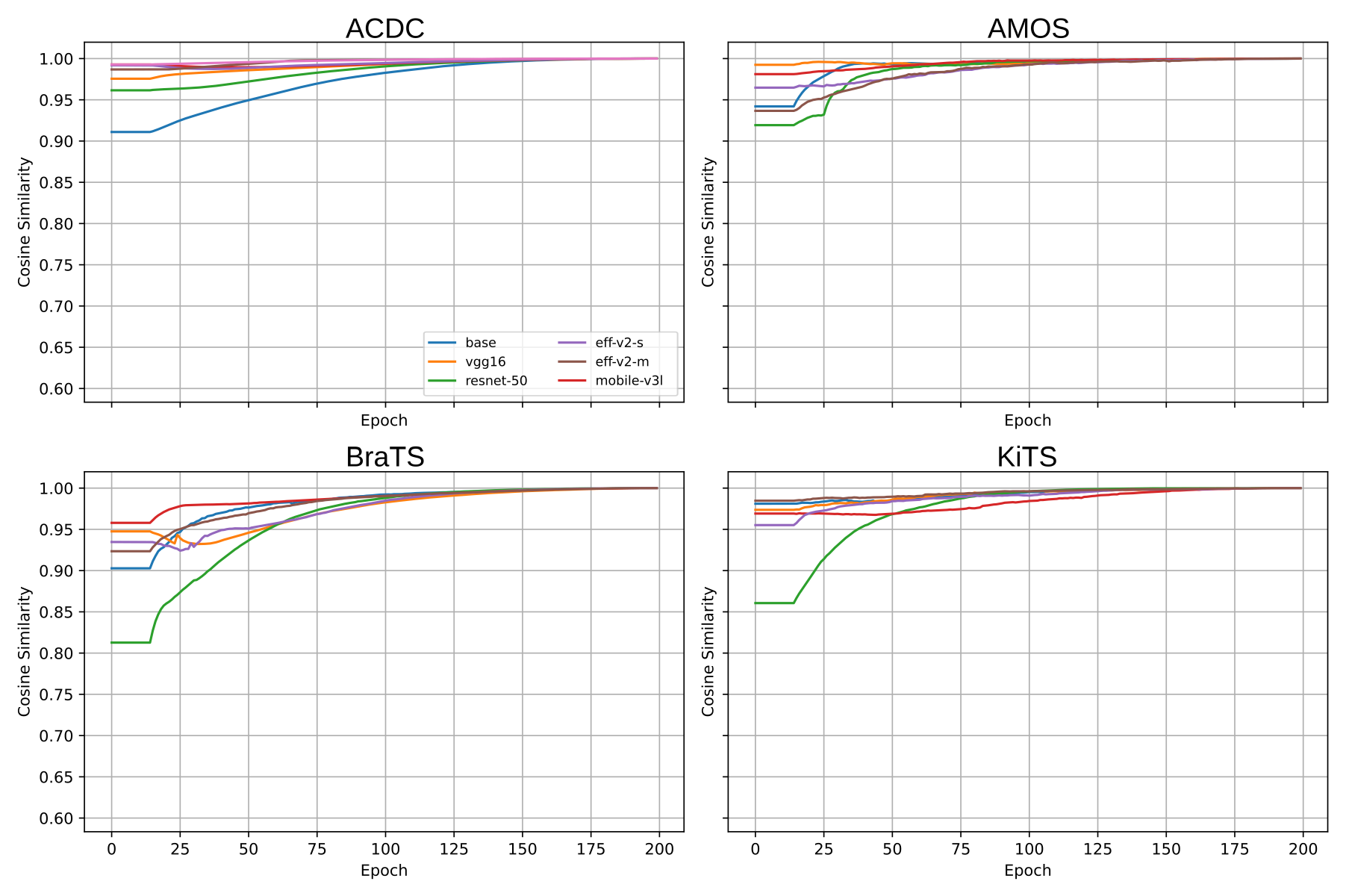}
	\caption{Global convergence of architecture parameters.
		Cosine similarity between the current $\alpha \cdot \beta$ vector and its final value is plotted over 200 epochs for four datasets (ACDC, AMOS, BraTS, KiTS). 
		Each colored curve corresponds to one U-Net encoder variant (base, VGG-16, ResNet-50, MobileNetV3-Large, EfficientNetV2-S/M).}
	\label{fig:5}
\end{figure*}

Figure~\ref{fig:5} summarizes this behavior more compactly by plotting cosine similarity between $\boldsymbol{\Theta}^{(e)}$ and $\boldsymbol{\Theta}^{(E)}$ (final epoch) across epochs $e$. 
Across all datasets and backbones, similarity rises steeply in the first 40-60 epochs and exceeds $0.95$ around epoch~100, after which it asymptotically approaches~1.0. 
Larger backbones (e.g.\ ResNet-50 on KiTS) converge slightly more slowly, but the qualitative pattern is consistent: the continuous architecture parameters effectively stabilize by mid-search.

\begin{figure*}[ht!]
	\centering
	\includegraphics[width=0.8\textwidth]{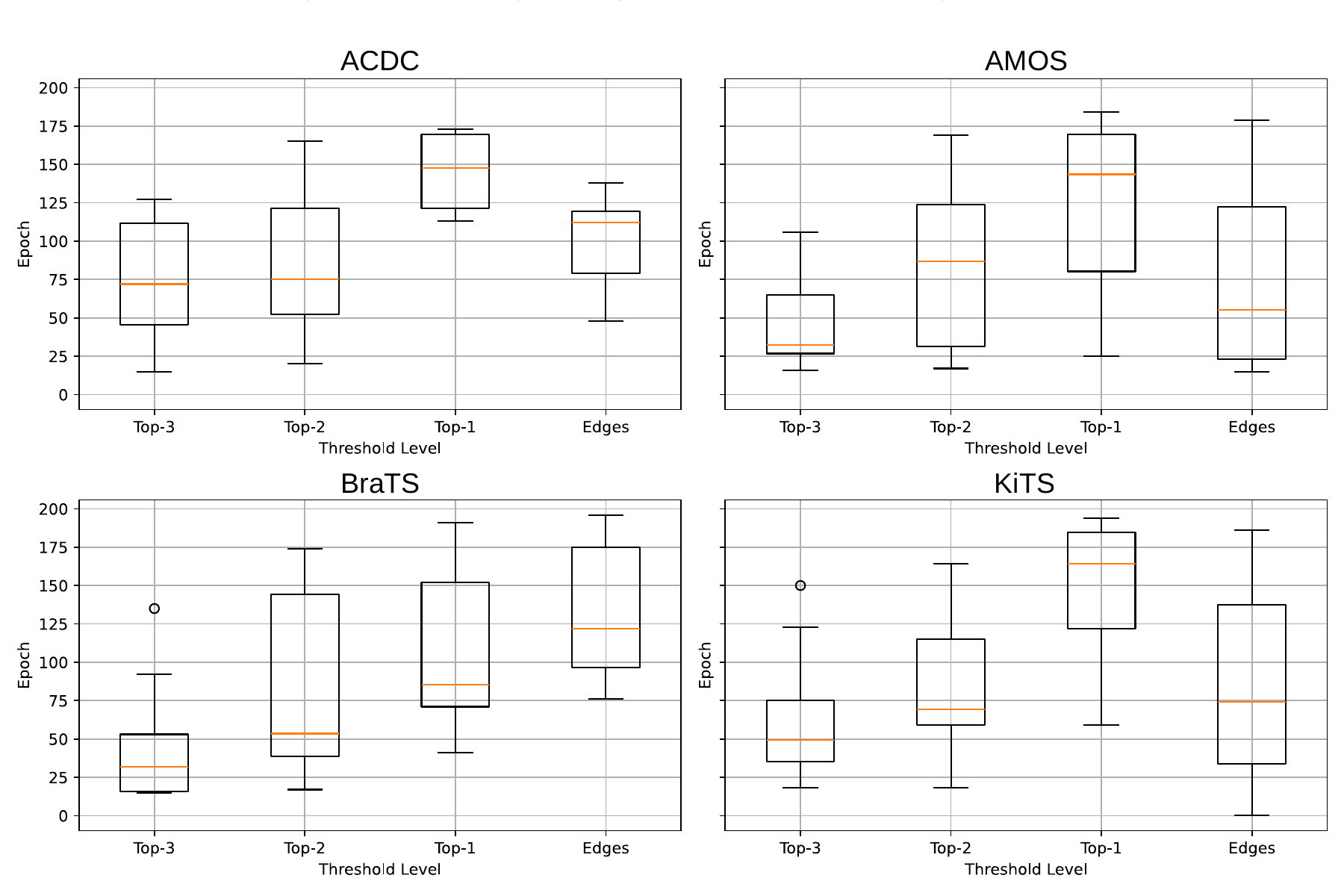}
	\caption{When does the final genotype first emerge during search?
		For each dataset (ACDC, AMOS, BraTS, KiTS) we record, across all U-Net backbones, the epoch at which the operation (or edge) that ends up in the final discrete cell first meets four threshold levels: 
		\emph{Top-3} -- appears among the three highest-scoring operations on its edge; 
		\emph{Top-2} -- appears among the two highest; 
		\emph{Top-1} -- becomes the single highest; 
		\emph{Edges} -- the two winning incoming edges of every node are fixed. 
		Each box-and-whisker plot summarizes the distribution of those epochs over all edges in the cell and all random seeds; medians are marked in orange.}
	\label{fig:6}
\end{figure*}

If we require full discrete alignment with the final architecture, however, nearly the entire budget is needed. 
Figure~\ref{fig:6} reports, for each dataset, the epoch at which the eventually selected operation or edge first satisfies four criteria: being in the top-3, top-2, top-1 on its edge, and forming the final set of incoming edges. 
Median epochs show that the final-winning operation typically enters the top-3 by approximately 45 epochs, reaches top-2 around 140 epochs, and achieves full top-1 and edge alignment only close to epochs 170-190. 
On more informative datasets such as BraTS, these thresholds are reached earlier.

\subsection{Edge-wise difficulty and environment effects}

\begin{figure*}[ht!]
	\centering
	\includegraphics[width=0.8\textwidth]{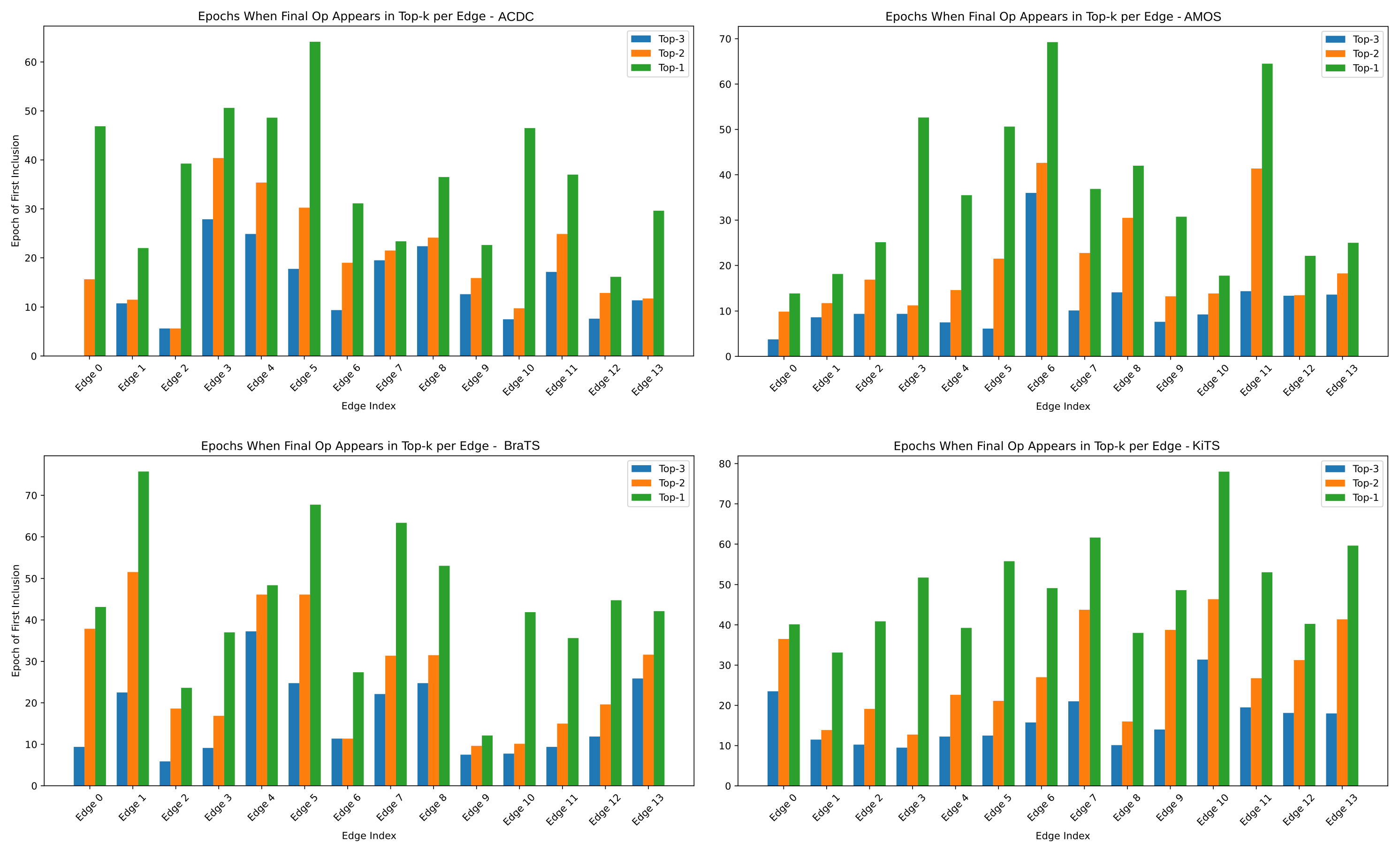}
	\caption{Per-edge difficulty of convergence.
		For each dataset (ACDC, AMOS, BraTS, KiTS) we chart, for every edge of the IAC cell, the epoch at which the operation that ends up in the final discrete architecture first appears in the Top-3 (blue), Top-2 (orange) or Top-1 (green) set on that edge. 
		Edges whose green bars emerge early are ``easy'' ones: the correct operation is evident right after the warm-up. 
		Edges with late-rising green bars are ``hard'' ones: they delay the moment when the whole cell is frozen. 
		Despite this variation, the correct operation is present on every edge by approximately 70 epochs in all datasets, confirming that individual operations converge quickly, while the subsequent assembly of edges and information flow is refined later to match the already learned backbone representations.}
	\label{fig:7}
\end{figure*}

Figure~\ref{fig:7} breaks down the emergence of final-winning operations \emph{per edge}. 
We observe a clear split between ``easy'' edges, where the correct operation becomes top-1 almost immediately after warm-up, and ``hard'' edges, which take much longer to settle. 
Intuitively, the more hard edges a cell has, the later its final discrete topology is fixed.

Despite this variation, the correct operation is present in the top-3 on \emph{every} edge by approximately epoch~70 across all datasets. 
This indicates that individual operations converge quickly and that later epochs are mainly used to refine how those operations are assembled into a coherent information flow through the DAG. 
This late-stage refinement depends not only on the IAC weights $\omega$ but also on the pre-trained backbone environment $\omega_U$: changes to the backbone or to the random initialization of $\omega$ can lead to different final architectures, consistent with observations from the Lottery Ticket Hypothesis in other settings.

\subsection{Summary of analysis}

The above results support three key observations:

\begin{itemize}
	\item The IAC architecture evolves rapidly at the beginning of the search, but the continuous architecture parameters stabilize much earlier (around 40-100 epochs) than the discrete top-1 decisions.
	\item Final-winning operations typically appear among the top-3 candidates on their edges very early in the search (within approximately 45 epochs) and remain in this small set until discretization.
	\item Every edge has identified its final operation among the top-3 by around epoch~70, even though selecting the exact top-1 and the final set of incoming edges may require most of the original budget.
\end{itemize}

These findings suggest that the search space can be reduced aggressively once per-edge operation-importance distributions stabilize, without sacrificing the quality of the final cell. 
In the next section we turn these insights into a JS-divergence-based pruning scheme that accelerates IAC search by several times (up to an order of magnitude in wall-clock time) while preserving segmentation performance.

\section{Proposed Approach}
\label{sec:method}

The analysis in Section~\ref{sec:analysis} shows that operation importances inside the IAC cell stabilize much earlier than the final discrete genotype and that the operations selected in the final discrete cell appear among the top candidates very early. 
We therefore propose an accelerated IAC search that exploits this behavior: we monitor per-edge operation-importance distributions over epochs using the Jensen--Shannon divergence and prune low-importance operations once these distributions have stabilised. 
The backbone pre-training and final IAC training stages remain unchanged; only the search phase of the original IAC pipeline is modified.
We refer to the resulting accelerated framework as \emph{IAC-LTH}.

Figure~\ref{fig:schema} provides a high-level overview, and Algorithm~\ref{algs:alg1} gives pseudocode of the proposed search procedure.

\subsection{Warm-up and joint optimization of $(\omega,\alpha,\beta)$}

We keep the warm-up strategy of the baseline IAC framework with a small modification. 
For the first $\xi = 15$ epochs we update only the IAC weights $\omega$ while keeping the architecture parameters $(\alpha,\beta)$ fixed (initialized to constant values). 
This allows the candidate operations to adapt to the frozen backbone $\omega_U$ before architecture selection begins, mitigating instabilities caused by simultaneous optimization of $\omega$ and $(\alpha,\beta)$ against a fixed environment.

After warm-up, each epoch of the search phase consists of standard training updates on the supernet:
\begin{equation}
	\min_{\omega,\alpha,\beta} \; \mathcal{L}_{\mathrm{train}}(\omega, \alpha, \beta, \omega_U).
	\label{eq:joint_train_proposed}
\end{equation}

Unlike the original IAC search, which alternates between updating $\omega$ on a dedicated \textit{train\_search} subset and updating $(\alpha,\beta)$ on a separate \textit{val\_search} subset (Section~\ref{sec:baseline_bilevel}), IAC-LTH uses a single-level optimization scheme inspired by SingleDARTS~\cite{bib:SingleDARTS}: 
$\omega$ is updated by SGD and $(\alpha,\beta)$ are updated by Adam on the \emph{same full training set}, without an internal \textit{train\_search}/\textit{val\_search} split.
This removes the inner validation loop and lets the IAC topology and its weights evolve jointly on all available training data.

\subsection{Operation importance and per-edge distributions}

For each edge $(i,j)$ and candidate operation $o \in O$ we define the importance at epoch $e$ as
\begin{equation}
	\Theta^{o,(e)}_{i,j} = \alpha^{o,(e)}_{i,j} \cdot \beta^{(e)}_{i,j},
	\label{eq:theta_def}
\end{equation}
consistent with the analysis in Section~\ref{sec:analysis_genotypes}.
Collecting all $\Theta^{o,(e)}_{i,j}$ yields the continuous genotype vector $\boldsymbol{\Theta}^{(e)}$ used in Figures~\ref{fig:3}-\ref{fig:5}.

To compare successive epochs on a per-edge basis we normalize $\Theta^{o,(e)}_{i,j}$ into a categorical distribution over operations:
\begin{equation}
	p^{(e)}_{i,j}(o) = 
	\frac{\Theta^{o,(e)}_{i,j}}{\sum\nolimits_{o' \in O} \Theta^{o',(e)}_{i,j} + \varepsilon},
	\label{eq:edge_dist}
\end{equation}
where $\varepsilon$ is a small constant to avoid division by zero (if the denominator is zero, we treat $p^{(e)}_{i,j}$ as uniform). 
This distribution reflects the relative preference for each operation on edge $(i,j)$ at epoch $e$.

At the end of every epoch of the search phase we recompute $\Theta^{o,(e)}_{i,j}$ and $p^{(e)}_{i,j}$ for all edges; these values are then used for stability tests and pruning.

\subsection{JS-divergence-based stability test}

To detect when an edge has stabilized, we compare its operation-importance distributions between successive epochs using the Jensen--Shannon divergence. 
For edge $(i,j)$ and epochs $e-1$ and $e$ we define
\begin{equation}
	D_{\mathrm{JS}}\big(p^{(e-1)}_{i,j} \Vert p^{(e)}_{i,j}\big)
	= \tfrac{1}{2} D_{\mathrm{KL}}\big(p^{(e-1)}_{i,j} \Vert m_{i,j}\big)
	+ \tfrac{1}{2} D_{\mathrm{KL}}\big(p^{(e)}_{i,j} \Vert m_{i,j}\big),
\end{equation}
where $m_{i,j} = \tfrac{1}{2}\big(p^{(e-1)}_{i,j} + p^{(e)}_{i,j}\big)$ and $D_{\mathrm{KL}}$ denotes Kullback--Leibler divergence. 
JS divergence is symmetric and bounded, which makes it convenient for threshold-based criteria.

For each edge $(i,j)$ we maintain an individual JS threshold $\theta_{i,j} > 0$ and a multiplicative factor $\kappa > 1$. 
At epoch $e$ we apply the following rule:

\begin{itemize}
	\item If $D_{\mathrm{JS}}\big(p^{(e-1)}_{i,j} \Vert p^{(e)}_{i,j}\big) < \theta_{i,j}$, we regard edge $(i,j)$ as \emph{stable} and trigger pruning of low-importance operations on that edge.
	\item Otherwise we increase the threshold, $\theta_{i,j} \leftarrow \kappa \theta_{i,j}$, making the stability criterion progressively easier to satisfy on edges whose distributions keep changing.
\end{itemize}

This mechanism allows early pruning on ``easy'' edges, whose importance distributions quickly stabilize, while letting ``hard'' edges continue to evolve until their JS divergence has decreased sufficiently.

\subsection{Pruning operations on stable edges}

When edge $(i,j)$ passes the JS stability test at epoch $e$, we prune low-importance operations according to the current importance $\Theta^{o,(e)}_{i,j}$.
We first compute the mean and standard deviation over operations on that edge:
\begin{equation}
	\mu_{i,j} = \frac{1}{|O|} \sum_{o \in O} \Theta^{o,(e)}_{i,j}, 
	\qquad
	\sigma_{i,j} = \sqrt{\frac{1}{|O|} \sum_{o \in O} \big(\Theta^{o,(e)}_{i,j} - \mu_{i,j}\big)^2}.
\end{equation}

We then apply a two-stage pruning rule:

\begin{enumerate}
	\item \textbf{Outlier pruning:}  
	prune all operations $o$ such that $\Theta^{o,(e)}_{i,j} < \mu_{i,j} - 2\sigma_{i,j}$ (strong negative outliers).
	\item \textbf{Fallback pruning:}  
	if no outliers are found, prune exactly one operation: the operation with the smallest importance $\Theta^{o,(e)}_{i,j}$ on that edge.
\end{enumerate}

We ensure that at least one operation always remains active on each edge: if all operations on an edge would be removed, we retain the operation with the largest importance. 
In epochs where the JS condition is not met, no operation is pruned on that edge and only the threshold $\theta_{i,j}$ is updated.

\subsection{Edge pruning and early stopping}

As operations are pruned, each edge $(i,j)$ eventually retains only a single active operation. 
At this point the mixed operation degenerates into a single operator. 
We continue the search until every edge in the IAC cell satisfies this condition; this corresponds to full discretization at the operation level.

In the final step we prune edges themselves to enforce the standard IAC connectivity pattern: for each node $j$ we keep only the two incoming edges with the largest edge weights $\Psi_{i,j}$ (equivalently, the largest $\beta_{i,j}$ values) and remove all other incoming edges, as in the baseline discretization (Section~\ref{sec:baseline_discretization}). 
The resulting sparse DAG defines the final discrete IAC genotype.

Once all edges have a single operation and each node has at most two incoming edges, the architecture is fully specified and further optimization of $(\alpha,\beta)$ is unnecessary. 

We therefore terminate the search phase \emph{early} (often after a fraction of the original 200-epoch budget) and proceed directly to the final IAC training stage, reusing the discovered genotype and initializing the IAC weights $\omega$ from their values at the beginning of the search. 
During this stage the cell is trained at full capacity (all channels, without partial-channel masks), exactly as in the original IAC framework.

\subsection{Algorithm}
\label{sec:method_algo}

Algorithm~\ref{algs:alg1} summarizes the proposed accelerated search. 
Compared to the baseline IAC search, the main additions are: 
(i) computation of per-edge operation-importance distributions, 
(ii) JS-divergence-based stability tests, and 
(iii) progressive pruning of low-importance operations on stable edges and, in the final step, reduces incoming edges per node to the two most influential ones, matching the original IAC connectivity pattern;

\begin{algorithm}[H]
	\scriptsize
	\caption{\small{Accelerated IAC Search (IAC-LTH)}\\
		\noindent
		\scriptsize
		\fcolorbox[HTML]{fae6e5}{fae6e5}{Initialization}\quad
		\fcolorbox[HTML]{fef2e6}{fef2e6}{Stopping criterion}\quad
		\fcolorbox[HTML]{ecf3fc}{ecf3fc}{Warm-up}\quad
		\fcolorbox[HTML]{eaf3e9}{eaf3e9}{Optimization}\quad
		\fcolorbox[HTML]{efeaf2}{efeaf2}{Importance / JS}\quad
		\fcolorbox[HTML]{fef8e6}{fef8e6}{Pruning}}
	\label{algs:alg1}
	\begin{algorithmic}[1]
		\Require Architecture parameters $\alpha,\beta$; IAC weights $\omega$; frozen U-Net weights $\omega_U$; warm-up epochs $\xi$; training set $D_{\text{train}}$ (with $\zeta$ steps per epoch); loss $L$; optimizers $\hat{\eta}$ (for $\omega$) and $\hat{\iota}$ (for $\alpha,\beta$); initial per-edge thresholds $\theta_{i,j}$; threshold multiplier $\kappa$.
		
		\State \vspace{-1pt}\fcolorbox[HTML]{fae6e5}{fae6e5}{\strut \parbox{0.99\linewidth}{\textbf{Init:} Build supernet $S$ with IAC cell, load $\omega_U$. Set $\alpha,\beta \gets \text{const}$. Randomly initialize $\omega$. Freeze all weights except $\{\alpha,\beta,\omega\}$.}}
		
		\State \vspace{-1pt}\fcolorbox[HTML]{fef2e6}{fef2e6}{\strut \parbox{0.99\linewidth}{\textbf{Stop:} Every IAC edge has exactly one active operation (one op per edge).}}
		
		\State \vspace{-1pt}\fcolorbox[HTML]{ecf3fc}{ecf3fc}{\strut \parbox{0.99\linewidth}{\textbf{Warm-up:} \textbf{for} $e = 1$ \textbf{to} $\xi$ \textbf{do}}}
		\State \vspace{-1pt}\fcolorbox[HTML]{ecf3fc}{ecf3fc}{\strut \parbox{0.99\linewidth}{\hspace{1em}\textbf{for} $z = 1$ \textbf{to} $\zeta$ \textbf{do}}}
		\State \vspace{-1pt}\fcolorbox[HTML]{ecf3fc}{ecf3fc}{\strut \parbox{0.99\linewidth}{\hspace{2em}Sample a mini-batch from $D_{\text{train}}$ and compute $L_{\text{train}} = L(S(D_{\text{train}}))$.}}
		\State \vspace{-1pt}\fcolorbox[HTML]{ecf3fc}{ecf3fc}{\strut \parbox{0.99\linewidth}{\hspace{2em}$\omega \gets \omega - \hat{\eta}\,\nabla_{\omega} L_{\text{train}}(\omega,\alpha,\beta,\omega_U)$.}}
		
		\State \textbf{while} stopping criterion not met \textbf{do}
		\State \vspace{-1pt}\fcolorbox[HTML]{eaf3e9}{eaf3e9}{\strut \parbox{0.99\linewidth}{\hspace{1em}\textbf{for} $z = 1$ \textbf{to} $\zeta$ \textbf{do}}}
		\State \vspace{-1pt}\fcolorbox[HTML]{eaf3e9}{eaf3e9}{\strut \parbox{0.99\linewidth}{\hspace{2em}Sample a mini-batch and compute $L_{\text{train}} = L(S(D_{\text{train}}))$.}}
		\State \vspace{-1pt}\fcolorbox[HTML]{eaf3e9}{eaf3e9}{\strut \parbox{0.99\linewidth}{\hspace{2em}$\omega \gets \omega - \hat{\eta}\,\nabla_{\omega} L_{\text{train}}(\omega,\alpha,\beta,\omega_U)$.}}
		\State \vspace{-1pt}\fcolorbox[HTML]{eaf3e9}{eaf3e9}{\strut \parbox{0.99\linewidth}{\hspace{2em}$(\alpha,\beta) \gets (\alpha,\beta) - \hat{\iota}\,\nabla_{(\alpha,\beta)} L_{\text{train}}(\omega,\alpha,\beta,\omega_U)$.}}
		
		\State \vspace{-1pt}\fcolorbox[HTML]{efeaf2}{efeaf2}{\strut \parbox{0.99\linewidth}{\hspace{1em}\textbf{for each} edge $(i,j) \in E$ \textbf{do} \textit{/* importance and JS */}}}
		\State \vspace{-1pt}\fcolorbox[HTML]{efeaf2}{efeaf2}{\strut \parbox{0.99\linewidth}{\hspace{2em}Compute $\Theta^{o}_{i,j} = \alpha^{o}_{i,j} \cdot \beta_{i,j}$ for all $o \in O$ (current epoch).}}
		\State \vspace{-1pt}\fcolorbox[HTML]{efeaf2}{efeaf2}{\strut \parbox{0.99\linewidth}{\hspace{2em}Normalize to $p_{i,j}(o) = \Theta^{o}_{i,j}\big/\big(\sum_{o'}\Theta^{o'}_{i,j} + \varepsilon\big)$. Let $p'_{i,j}$ be the distribution from the previous epoch (or uniform in the first epoch).}}
		\State \vspace{-1pt}\fcolorbox[HTML]{efeaf2}{efeaf2}{\strut \parbox{0.99\linewidth}{\hspace{2em}Compute $d_{i,j} = D_{\mathrm{JS}}\!\big(p'_{i,j} \Vert p_{i,j}\big)$.}}
		
		\State \vspace{-1pt}\fcolorbox[HTML]{fef8e6}{fef8e6}{\strut \parbox{0.99\linewidth}{\hspace{1em}\textbf{for each} edge $(i,j) \in E$ \textbf{do} \textit{/* prune ops or adjust threshold */}}}
		\State \vspace{-1pt}\fcolorbox[HTML]{fef8e6}{fef8e6}{\strut \parbox{0.99\linewidth}{\hspace{2em}\textbf{if} $d_{i,j} < \theta_{i,j}$ \textbf{then}}}
		\State \vspace{-1pt}\fcolorbox[HTML]{fef8e6}{fef8e6}{\strut \parbox{0.99\linewidth}{\hspace{3em}Compute $\mu_{i,j}$ and $\sigma_{i,j}$ over $\{\Theta^{o}_{i,j}\}_{o\in O}$.}}
		\State \vspace{-1pt}\fcolorbox[HTML]{fef8e6}{fef8e6}{\strut \parbox{0.99\linewidth}{\hspace{3em}Prune all operations with $\Theta^{o}_{i,j} < \mu_{i,j} - 2\sigma_{i,j}$.}}
		\State \vspace{-1pt}\fcolorbox[HTML]{fef8e6}{fef8e6}{\strut \parbox{0.99\linewidth}{\hspace{3em}\textbf{if} no op was pruned \textbf{then} prune the op with minimal $\Theta^{o}_{i,j}$ (ensuring at least one op remains).}}
		\State \vspace{-1pt}\fcolorbox[HTML]{fef8e6}{fef8e6}{\strut \parbox{0.99\linewidth}{\hspace{2em}\textbf{else} $\theta_{i,j} \gets \kappa \,\theta_{i,j}$.}}
		
		\State \textbf{end while}
		
		\State \vspace{-1pt}\fcolorbox[HTML]{fef8e6}{fef8e6}{\strut \parbox{0.99\linewidth}{\textbf{Final edge reduction:} For each node $j$ keep only the two incoming edges $(i,j)$ with the largest $\Psi_{i,j}$ (equivalently, the largest $\beta_{i,j}$), dropping all other incoming edges.}}
	\end{algorithmic}
\end{algorithm}

\textbf{Hyperparameters for JS-based pruning:} For all experiments, we initialize the JS thresholds $\theta_{i,j}$ to $1.0$ and use the same initial value across all edges.
Whenever an edge fails the stability test in a given epoch, we update its threshold according to $\theta_{i,j} \gets \kappa \,\theta_{i,j}$ with $\kappa = 2$, which gradually relaxes the stability requirement for that edge.
For numerical stability, we compute all JS divergences with a small constant $\varepsilon = 10^{-6}$ added to the smoothed probabilities.
We do not impose any explicit upper or lower bounds on $\theta_{i,j}$; in practice, the combination of the $\mu_{i,j} - 2\sigma_{i,j}$ pruning rule and the finite number of pruning events keeps $\theta_{i,j}$ in a reasonable range.
The values of $\theta_{i,j}$, $\kappa$, and $\varepsilon$ were chosen empirically based on preliminary experiments and then kept fixed for all datasets, backbones and experimental regimes, so the schedule should be regarded as a simple heuristic rather than a heavily tuned hyperparameter set.

\section{Experiments and Results}
\label{sec:experiments}

In this section we evaluate the proposed accelerated IAC search on the same segmentation benchmarks as the original IAC work and add experiments in the nnU-Net pipeline. 
Our goals are to verify that:
(i) IACs found by the accelerated search match the segmentation quality of the original full IAC search across multiple U-Net backbones; 
(ii) the improvements over plain skip connections are statistically significant at the patient level; and 
(iii) the LTH-inspired pruning strategy reduces the effective search budget by several times without sacrificing accuracy.
Unless stated otherwise, the Dice similarity coefficient is reported on held-out test patients.

\subsection{Datasets}

We use four public medical image segmentation benchmarks:
\begin{itemize}
	\item \textbf{ACDC} -- short-axis cardiac cine MRI with left ventricular myocardium and cavity segmentation.
	\item \textbf{AMOS} -- multi-organ abdominal CT with multiple soft-tissue structures.
	\item \textbf{BraTS} -- multi-modal brain MRI with tumor sub-structures.
	\item \textbf{KiTS} -- abdominal CT with kidney and tumor labels.
\end{itemize}

For each dataset we follow a patient-level split into training, validation and test subsets with disjoint patients.
The test sets contain 60, 72, 250, and 98 cases (patients) for ACDC, AMOS, BraTS and KiTS, respectively, shared across all backbones on a given dataset.
All methods use the same 2-D slice protocol, reflecting a practically common setting for medical segmentation pipelines.
We extract axial slices from each 3-D volume and apply a fixed $128\times128$ in-plane centered crop to reduce background and standardize the input size.
For MRI (ACDC, BraTS) we perform per-modality $z$-score normalization over non-zero voxels; for CT (AMOS, KiTS) we apply standard CT windowing followed by $z$-score normalization.
The same pre-processing is used for all backbones and skip variants.

\subsection{Architectures and baselines}

We consider two families of segmentation backbones.

\textbf{Frozen U-Nets with different encoders (backbone grid):} We reuse the six encoder variants from the original IAC study: a plain U-Net encoder plus several modern CNN backbones.
For the backbone-grid experiments, we use a single random seed per (dataset, backbone) pair; the reported statistics are averages over backbones only.
For each backbone we compare:
\begin{itemize}
	\item \textbf{Ref} -- reference U-Net with standard identity skip connections.
	\item \textbf{Att} -- U-Net with Attention Gates.
	\item \textbf{Nested} -- U-Net++-style dense skip connections.
	\item \textbf{ResAtt} -- U-Net with residual attention blocks.
	\item \textbf{IAC} -- original IAC obtained with a full 200-epoch differentiable search.
	\item \textbf{IAC-LTH} -- the proposed accelerated IAC search (same search space and backbone, but with operation-level analysis, JS-based stability detection and on-the-fly pruning).
\end{itemize}

\textbf{nnU-Net pipeline:} We further integrate IACs into the standard 2-D nnU-Net framework and repeat the experiments both without and with data augmentation.
All nnU-Net search runs were performed with three different random seeds per dataset and method (IAC and IAC-LTH), for both the non-augmented and augmented regimes, and we report averages over these runs:
\begin{itemize}
	\item \textbf{nnU-Net (Ref)} -- unmodified 2-D nnU-Net.
	\item \textbf{nnU-Net + IAC} -- skip connections replaced with IACs found by the original 200-epoch search.
	\item \textbf{nnU-Net + IAC-LTH} -- skip connections replaced with IACs found by the proposed accelerated search.
\end{itemize}

All IAC variants use the same cell topology, search space and final training protocol described in Section~\ref{sec:baseline_iac} and Section~\ref{sec:method}.

\subsection{Training protocol}
\label{sec:experiments_setup}

All models use loss functions that follow nnU-Net conventions \cite{bib:nnUNet, bib:nnUNet2024}:
\begin{equation}
	L=\lambda\,(1-\overline{D}_{\text{soft}})+ (1-\lambda)\,L_{\text{CE}},
\end{equation}
where
\begin{equation}
	\overline{D}_{\text{soft}}=\frac{1}{|\mathcal F|}\sum_{c\in\mathcal F}
	\frac{2\sum_x y_c p_c+\varepsilon}{\sum_x y_c+\sum_x p_c+\varepsilon}
\end{equation}
is the macro-average soft Dice over foreground classes (background excluded),
\begin{equation}
	L_{\text{CE}}=-\frac{1}{|\Omega|}\sum_x\sum_c y_c\log p_c
\end{equation}
with $\lambda=0.5$ and $\varepsilon=10^{-5}$.
Here, $y_c$ denotes the one-hot ground-truth for class $c$, $p_c$ the softmax probability, $\Omega$ the set of pixels, and $\mathcal{F}$ the set of foreground classes.
When deep supervision is enabled (nnU-Net setting), we sum the loss over all decoder outputs using the standard nnU-Net output weights.
For the backbone-grid experiments we disable deep supervision to isolate architectural effects.

For evaluation, we compute Dice per test case (patient) for each foreground class and report per-class means and the mean foreground Dice (macro-average over foreground classes, background excluded).
Dataset scores are averages of these per-case/patient-level values.
Although all models operate on 2-D slices, evaluation is performed at the case level by aggregating true positives (TP), false positives (FP) and false negatives (FN) over all slices of a patient before computing Dice.

\textbf{Stage I (reference backbones).} Each 2-D U-Net backbone is trained for 200 epochs using Adam (learning rate 0.001, batch size 128), selecting the best checkpoint on the validation set based on mean patient-level foreground Dice.

\textbf{Stage III (discrete cells).} After search and discretization, we instantiate the discrete IAC / IAC-LTH cell in the U-Net that was used during search and train only the IAC / IAC-LTH weights for 200 epochs with the same optimizer, learning rate and loss as in Stage~I, while keeping the backbone frozen.
For the baseline IAC we follow the original regime and \emph{re-initialize} the IAC weights with a fresh random initialization before this final training stage.
For IAC-LTH, following the Lottery Ticket protocol, we instead \emph{reset} the weights of the discovered subnetwork to the values they had at the start of the search (i.e., the \emph{initial} random initialization stored before any updates) and train from this preserved initialization.

For the backbone grid and nnU-Net, we follow the original IAC setup for the search phase:
\begin{itemize}
	\item Search is performed once per backbone and dataset with frozen encoder-decoder weights.
	\item For the baseline IAC, we use the original split of the training data into \textit{train\_search} and \textit{val\_search} and alternate updates of $\omega$ and $(\alpha,\beta)$ on these subsets, as in Section~\ref{sec:baseline_bilevel}.
	\item For IAC-LTH, we use the same overall training set but do \emph{not} create an internal validation subset: $\omega$ and $(\alpha,\beta)$ are optimized jointly on the full training set according to Eq.~\eqref{eq:joint_train_proposed}.
	\item The IAC search phase runs for up to 200 epochs in the full-search baseline; the accelerated variant applies JS-based stability checks and pruning after each epoch and thus typically terminates much earlier (Section~\ref{sec:search_efficiency}).
\end{itemize}
Unless stated otherwise, both IAC and IAC-LTH use the same supernet optimization hyperparameters (SGD for cell weights with learning rate 0.01, Adam for architecture parameters with learning rate 0.001, cosine power annealing, batch size 128); only the use of the internal train/validation split and the JS-based pruning differ.

\subsection{Backbone grid: Dice gains over reference U-Net}
\label{sec:backbone_grid_results}

We first evaluate how different skip-connection refinements behave across the grid of six U-Net backbones. 
For each dataset and method we compute the per-backbone patient-level Dice difference relative to the reference U-Net,
\(
\Delta\text{Dice} = \text{Dice}_{\text{method}} - \text{Dice}_{\text{Ref}},
\)
and then average over the six backbones.

Table~\ref{tab:backbone_delta} summarizes the mean $\Delta\text{Dice}$ for all skip variants: Attention, Nested, Residual Attention, IAC and IAC-LTH, together with 95\% bootstrap confidence intervals. 
All numbers are expressed as absolute Dice differences.

Unless stated otherwise, statistical significance is assessed at the patient level.
For pairwise method comparisons on a given dataset and backbone we use the Wilcoxon signed-rank test on per-patient Dice scores.
To summarize significance across backbones, we aggregate the resulting $p$-values using Fisher's method:
\[
\chi^2 = - 2 \sum_{k=1}^{K} \ln p_k, \qquad \chi^2 \sim \chi^2_{2K},
\]
where $p_k$ is the $p$-value for backbone $k$ and $K$ is the number of backbones.
We report the corresponding combined $p$-value as $p_{\text{Fisher}}$ in Table~\ref{tab:backbone_delta}.

\begin{table}[t]
	\centering
	\footnotesize
	\setlength{\tabcolsep}{4pt}
	\caption{Backbone grid: mean Dice difference vs.\ reference U-Net ($\Delta\text{Dice} = \text{Dice}_{\text{method}} - \text{Dice}_{\text{Ref}}$) across six backbones. Values in brackets are 95\% bootstrap confidence intervals over backbones. The last column reports global patient-level significance $p_{\text{Fisher}}$ (Fisher aggregation over backbones and patients).}
	\label{tab:backbone_delta}
	\begin{tabular}{llcc}
		\toprule
		Dataset & Method & $\Delta$Dice & $p_{\text{Fisher}}$ \\
		\midrule
		ACDC & Attention      & 0.068 [\,-0.000, 0.192] & $4.1\times 10^{-19}$ \\
		ACDC & Nested         & 0.034 [0.013, 0.065]    & $7.3\times 10^{-39}$ \\
		ACDC & Residual Att.  & 0.048 [0.002, 0.128]    & $3.0\times 10^{-21}$ \\
		ACDC & IAC            & 0.011 [0.000, 0.025]    & $1.7\times 10^{-25}$ \\
		ACDC & IAC-LTH        & 0.010 [0.001, 0.020]    & $4.5\times 10^{-15}$ \\
		\midrule
		AMOS & Attention      & -0.063 [-0.165, 0.027]  & $1.6\times 10^{-61}$ \\
		AMOS & Nested         & 0.031 [\,-0.043, 0.106] & $3.1\times 10^{-53}$ \\
		AMOS & Residual Att.  & 0.055 [\,-0.011, 0.129] & $2.5\times 10^{-51}$ \\
		AMOS & IAC            & 0.055 [\,-0.041, 0.134] & $1.2\times 10^{-47}$ \\
		AMOS & IAC-LTH        & 0.050 [\,-0.029, 0.125] & $3.2\times 10^{-57}$ \\
		\midrule
		BraTS & Attention     & 0.008 [0.003, 0.013]    & $1.5\times 10^{-30}$ \\
		BraTS & Nested        & 0.010 [0.005, 0.016]    & $1.8\times 10^{-61}$ \\
		BraTS & Residual Att. & 0.007 [0.002, 0.013]    & $6.7\times 10^{-34}$ \\
		BraTS & IAC           & 0.013 [0.007, 0.019]    & $1.1\times 10^{-67}$ \\
		BraTS & IAC-LTH       & 0.012 [0.006, 0.019]    & $5.1\times 10^{-59}$ \\
		\midrule
		KiTS & Attention      & -0.005 [-0.031, 0.020]  & $4.1\times 10^{-22}$ \\
		KiTS & Nested         & 0.012 [\,-0.016, 0.039] & $1.4\times 10^{-21}$ \\
		KiTS & Residual Att.  & 0.017 [\,-0.011, 0.047] & $1.7\times 10^{-24}$ \\
		KiTS & IAC            & 0.032 [0.005, 0.060]    & $1.2\times 10^{-29}$ \\
		KiTS & IAC-LTH        & 0.025 [0.004, 0.045]    & $1.0\times 10^{-25}$ \\
		\bottomrule
	\end{tabular}
\end{table}

Several trends emerge.

\begin{itemize}
	\item On \textbf{ACDC}, all refined skip variants yield positive mean $\Delta$Dice over the six backbones.  
	Attention has the largest average gain (but with a wide confidence interval that includes zero), Nested and Residual Attention are in the middle, while IAC and IAC-LTH deliver smaller but consistent gains with tight intervals just above zero.
	\item On \textbf{AMOS}, Attention slightly \emph{hurts} performance on average, whereas all other methods improve over the reference U-Net.  
	IAC and Residual Attention achieve the largest mean gains ($\approx 0.055$ Dice), with IAC-LTH close behind ($0.050$ Dice) and Nested somewhat lower.
	\item On \textbf{BraTS}, all skip variants provide small but consistently positive gains.  
	Here IAC and IAC-LTH attain the highest average improvements ($0.013$ and $0.012$ Dice, respectively), followed by Nested and Attention/Residual Attention.
	\item On \textbf{KiTS}, Attention is again neutral or slightly negative on average, while all other methods bring positive gains.  
	IAC achieves the largest mean improvement ($0.032$ Dice), IAC-LTH is the second-best ($0.025$ Dice), and Nested and Residual Attention lag behind, with confidence intervals that overlap zero.
\end{itemize}

The last column of Table~\ref{tab:backbone_delta} reports Fisher-aggregated patient-level $p$-values for IAC and IAC-LTH, combining all backbones and patients. 
All these $p_{\text{Fisher}}$ values are extremely small (between $10^{-15}$ and $10^{-67}$), confirming that the gains of IAC and IAC-LTH over the reference U-Net are statistically significant at the patient level and not driven by a few lucky backbones.

In summary, across the backbone grid the proposed IAC-LTH is consistently competitive with, and often stronger than, classical skip refinements such as Attention, Nested and Residual Attention:

\begin{itemize}
	\item on \textbf{AMOS} and \textbf{KiTS}, it outperforms Attention and Nested, and matches or slightly trails the best residual variant;
	\item on \textbf{BraTS}, IAC and IAC-LTH obtain the largest average gains among all skip types;
	\item on \textbf{ACDC}, all refined skips help, with IAC/IAC-LTH providing small but robust improvements with narrow confidence intervals.
\end{itemize}

Thus, within the family of plug-in skip modules for medical image segmentation, the accelerated IAC search preserves the behavior of IAC while remaining at least on par with, and in several cases better than, widely used alternatives such as attention-gated and nested skip connections.

\subsection{nnU-Net without data augmentation}
\label{sec:nnunet_noaug}

We next move to the 2-D nnU-Net baseline without any data augmentation. 
This creates a controlled but challenging setting in which the architecture has to compensate for lack of regularization and therefore tests whether the searched skip modules retain value in a strong medical segmentation pipeline.
For each dataset we compare:
\begin{itemize}
	\item nnU-Net (Ref),
	\item nnU-Net + IAC,
	\item nnU-Net + IAC-LTH,
\end{itemize}
and compute patient-level Dice differences between all pairs.

We define
\[
\Delta\text{Dice}(A\rightarrow B) = \text{Dice}_{B} - \text{Dice}_{A},
\]
so positive values mean that method $B$ improves over method $A$.
Table~\ref{tab:nnunet_noaug} summarizes the mean $\Delta\text{Dice}$, with 95\% bootstrap confidence intervals.

\begin{table*}[t]
	\centering
	\footnotesize
	\caption{nnU-Net without data augmentation: mean pairwise Dice differences $\Delta\text{Dice}(A\rightarrow B)$ at the patient level. Positive values indicate that method $B$ outperforms method $A$. Values in brackets: 95\% bootstrap confidence intervals.}
	\label{tab:nnunet_noaug}
	\begin{tabular}{lccc}
		\toprule
		Dataset & $\Delta$Dice(Ref$\rightarrow$IAC) & $\Delta$Dice(Ref$\rightarrow$IAC-LTH) & $\Delta$Dice(IAC$\rightarrow$IAC-LTH) \\
		\midrule
		ACDC & 0.023 [0.007, 0.042] & 0.025 [0.007, 0.045] & 0.002 [0.000, 0.003] \\
		AMOS & 0.367 [0.350, 0.385] & 0.363 [0.345, 0.381] & -0.004 [-0.008, 0.000] \\
		BraTS & 0.052 [0.043, 0.061] & 0.012 [0.003, 0.022] & -0.039 [-0.045, -0.034] \\
		KiTS & 0.161 [0.140, 0.184] & 0.182 [0.158, 0.208] & 0.021 [0.012, 0.031] \\
		\bottomrule
	\end{tabular}
\end{table*}

The first two columns confirm that both nnU-Net + IAC and nnU-Net + IAC-LTH improve over the unmodified nnU-Net on all four datasets. 
The gains are modest on ACDC and BraTS (about $+0.02$--$0.03$ Dice for IAC and IAC-LTH on ACDC, and about $+0.05$ Dice for IAC vs.\ $+0.01$ Dice for IAC-LTH on BraTS), and very large on the more challenging CT datasets AMOS and KiTS (up to $+0.37$ Dice without augmentation). 
All these improvements are statistically significant at the patient level (Wilcoxon $p \le 4.9\times 10^{-2}$, typically much smaller).

The third column shows the relative ranking of IAC vs.\ IAC-LTH in this regime. 
On ACDC, IAC-LTH is numerically almost identical to IAC (about $+0.002$ Dice; $p=3.5\times 10^{-11}$), i.e.\ the difference is statistically detectable but negligible in practice. 
On KiTS, IAC-LTH improves over IAC by about $+0.02$ Dice ($p=2.2\times 10^{-6}$). 
On AMOS, the difference is small and slightly negative ($-0.004$ Dice, $p=2.6\times 10^{-2}$), indicating that the two variants are essentially tied. 
On BraTS, IAC is ahead by about $0.04$ Dice ($p\approx 2.5\times 10^{-37}$). 
Overall, in the no-augmentation regime, the accelerated search retains the main improvements of IAC over nnU-Net, and the choice between full IAC and IAC-LTH has a second-order effect compared to their common gain over the baseline.

\subsection{nnU-Net with data augmentation}
\label{sec:nnunet_aug}

We now repeat the nnU-Net experiments with the full data augmentation pipeline enabled. 
This setting is closer to realistic medical segmentation training and tests whether the accelerated IAC search remains beneficial when the backbone is already strongly regularized.

Again we compute $\Delta\text{Dice}(A\rightarrow B)$ at the patient level. 
Table~\ref{tab:nnunet_aug} reports the mean differences and 95\% confidence intervals.

\begin{table*}[t]
	\centering
	\footnotesize
	\caption{nnU-Net with data augmentation: mean pairwise Dice differences $\Delta\text{Dice}(A\rightarrow B)$ at the patient level. Positive values indicate that method $B$ outperforms method $A$. Values in brackets: 95\% bootstrap confidence intervals.}
	\label{tab:nnunet_aug}
	\begin{tabular}{lccc}
		\toprule
		Dataset & $\Delta$Dice(Ref$\rightarrow$IAC) & $\Delta$Dice(Ref$\rightarrow$IAC-LTH) & $\Delta$Dice(IAC$\rightarrow$IAC-LTH) \\
		\midrule
		ACDC & 0.106 [0.082, 0.134] & 0.076 [0.055, 0.099] & -0.030 [-0.051, -0.014] \\
		AMOS & 0.145 [0.135, 0.154] & 0.448 [0.426, 0.470] & 0.303 [0.285, 0.321] \\
		BraTS & 0.048 [0.039, 0.056] & 0.042 [0.033, 0.051] & -0.006 [-0.014, 0.002] \\
		KiTS & 0.114 [0.099, 0.131] & 0.222 [0.192, 0.253] & 0.108 [0.085, 0.131] \\
		\bottomrule
	\end{tabular}
\end{table*}

The first two columns show that even in a strong augmented pipeline, inserting IACs into the skip connections yields consistent patient-level improvements on all datasets:
\begin{itemize}
	\item nnU-Net + IAC improves over nnU-Net by about $+0.11$ Dice on ACDC, $+0.15$ on AMOS, $+0.05$ on BraTS and $+0.11$ on KiTS (all $p \le 1.6\times 10^{-11}$).
	\item nnU-Net + IAC-LTH also improves over nnU-Net, with gains of $+0.08$ (ACDC), $+0.45$ (AMOS), $+0.04$ (BraTS) and $+0.22$ (KiTS), all with extremely small $p$-values ($p \le 1.7\times 10^{-13}$).
\end{itemize}

The third column reveals the interaction between data augmentation and the accelerated search. 
On ACDC, IAC is slightly ahead of IAC-LTH by about $0.03$ Dice ($p=7.0\times 10^{-5}$). 
On BraTS, the difference is very small and not statistically significant ($-0.006$ Dice, confidence interval includes zero, $p=0.13$), so the two variants can be considered equivalent. 
On the more challenging CT datasets AMOS and KiTS, however, IAC-LTH outperforms the full-search IAC: by $+0.30$ Dice on AMOS and $+0.11$ Dice on KiTS, both with very tight confidence intervals and $p \approx 1.7\times 10^{-13}$. 
In these regimes, the early-stabilized operation choices discovered by the accelerated search appear to interact favorably with strong data augmentation.

\subsection{Search efficiency and wall-clock cost}
\label{sec:search_efficiency}

While the Dice scores of IAC and IAC-LTH are closely matched, the main motivation for introducing the accelerated search is efficiency. 
In the original IAC framework, the differentiable search phase was always run for 200 epochs per backbone and dataset, regardless of whether the architecture had already stabilized. 
As shown in Section~\ref{sec:analysis}, the $\alpha\cdot\beta$ vectors converge much earlier: cosine similarity to the final configuration exceeds $0.95$ around epoch~100 and often stabilizes by 40-60 epochs, whereas the final discrete genotype only snaps into place near the end of the run.

The JS-based stability criterion in IAC-LTH exploits this behavior by monitoring, for each edge, the JS divergence between successive operation-importance distributions and pruning low-importance operations once these distributions stabilize. 
In practice this leads to a substantial reduction in both the number of search epochs and the wall-clock cost.

Table~\ref{tab:search_cost_backbones} reports the average search cost over the six U-Net backbones (Base, VGG-16, ResNet-50, MobileNetV3-Large, EfficientNetV2-S/M) for each dataset. 
The baseline IAC search always uses 200 epochs; IAC-LTH stops early once every edge has a single operation and each node has at most two incoming edges.

\begin{table*}[t]
	\centering
	\footnotesize
	\setlength{\tabcolsep}{4pt}
	\caption{Backbone grid: average search cost over six U-Net backbones for each dataset. IAC uses a fixed 200-epoch search. IAC-LTH stops early once the cell is fully discrete. Time is wall-clock GPU time for the search phase only. All times are in minutes, and the number of epochs is the average across multiple backbones for a given dataset.}
	\label{tab:search_cost_backbones}
	\begin{tabular}{lcccc}
		\toprule
		Dataset & IAC time & IAC-LTH epochs & IAC-LTH time & Time speed-up \\
		\midrule
		ACDC & 16.3 & 43 & 3.7  & $4.5\times$ \\
		AMOS & 632.7 & 48 & 124.3 & $5.1\times$ \\
		BraTS & 471.7 & 51 & 128.5 & $3.7\times$ \\
		KiTS & 234.5 & 48 & 53.7 & $4.4\times$ \\
		\bottomrule
	\end{tabular}
\end{table*}

Across the backbone grid, IAC-LTH consistently terminates after roughly 45-50 epochs instead of 200, i.e.\ about $4\times$ fewer epochs. 
This translates into a $3.7$-$5.1\times$ reduction in wall-clock search time, depending on the dataset, while preserving (and sometimes slightly improving) Dice compared to the full 200-epoch IAC search.

Table~\ref{tab:search_cost_nnunet} shows the same comparison inside the nnU-Net pipeline, with and without data augmentation. 
Again, the IAC search always runs for 200 epochs, whereas IAC-LTH stops after approximately 50-56 epochs.

\begin{table*}[t]
	\centering
	\footnotesize
	\setlength{\tabcolsep}{4pt}
	\caption{nnU-Net: search cost for IAC vs.\ IAC-LTH with and without data augmentation. IAC always runs for 200 epochs; IAC-LTH stops when the cell is fully discrete. Time is wall-clock GPU time for the search phase. All times are in minutes, and the number of epochs is the average across multiple seeds for a given dataset and configuration.}
	\label{tab:search_cost_nnunet}
	\begin{tabular}{llcccc}
		\toprule
		Dataset & Config & IAC time & IAC-LTH epochs & IAC-LTH time & Time speed-up \\
		\midrule
		ACDC & no aug & 37   & 48 & 11  & $3.4\times$ \\
		ACDC & aug    & 41   & 49 & 11  & $3.7\times$ \\
		AMOS & no aug & 631  & 56 & 179 & $3.5\times$ \\
		AMOS & aug    & 2635 & 55 & 162 & $16.3\times$ \\
		BraTS & no aug & 1110 & 54 & 401 & $2.8\times$ \\
		BraTS & aug    & 1162 & 57 & 409 & $2.8\times$ \\
		KiTS & no aug  & 515  & 54 & 179 & $2.9\times$ \\
		KiTS & aug     & 1009 & 51 & 163 & $6.2\times$ \\
		\bottomrule
	\end{tabular}
\end{table*}

In nnU-Net, the epoch reduction is again about $3.5$-$4\times$, and the wall-clock speed-up ranges from roughly $2.8\times$ (BraTS) to over $16\times$ in the heaviest configuration (AMOS with augmentation). 
The most extreme case, AMOS with nnU-Net and augmentation, drops from 2635 minutes ($\approx 44$ hours) for IAC to 162 minutes ($\approx 2.7$ hours) for IAC-LTH, while achieving higher Dice (Section~\ref{sec:nnunet_aug}).

Since both methods share the same deployed architecture size and inference pipeline, model FLOPs, parameter counts and test-time latency remain nearly identical, as in the original IAC paper. 
The main practical gain of IAC-LTH is therefore in \emph{offline search cost}, not runtime deployment: good IAC cells for medical segmentation can be discovered in a few dozen epochs and a couple of hours of GPU time instead of multi-hundred-epoch, multi-hour or even multi-day searches.

\subsection{Summary of empirical findings}

The experimental results can be distilled into three main messages:
\begin{enumerate}
	\item \textbf{IAC-LTH preserves, and in some settings enhances, the benefits of adaptive skip modules for medical segmentation.}  
	Across four datasets, multiple U-Net backbones and the nnU-Net pipeline, IAC-LTH delivers patient-level Dice gains very similar to or larger than those of the original IAC search. 
	In several CT scenarios (AMOS and KiTS with augmentation) it outperforms the full-search IAC.
	\item \textbf{Improvements are robust and statistically significant.}  
	Both backbone-grid and nnU-Net experiments show consistently positive $\Delta\text{Dice}$ vs.\ reference models, with very low $p$-values from Wilcoxon tests and Fisher-aggregated patient-level statistics for backbone grid experiments. 
	Gains are not limited to a few easy cases; they shift the entire patient distribution.
	\item \textbf{Search becomes cheaper without sacrificing accuracy.}  
	Thanks to early detection of ``winning'' operations and JS-based stability checks, the accelerated search shrinks the search space on the fly, prunes unpromising operations and terminates once the architecture has stabilized. 
	This reduces the search budget by several times (up to an order of magnitude in wall-clock time in the most demanding settings) while delivering cells whose segmentation performance is on par with, or in some settings better than, those produced by the full IAC procedure.
\end{enumerate}

These findings indicate that early stabilization inside Implantable Adaptive Cells can be exploited to make NAS more practical for medical image segmentation, preserving the accuracy benefits of adaptive skip modules while reducing the search cost.

\section{Discussion, limitations and positioning}

\textbf{Implications for medical image segmentation.}
The main result of this study is practical: adaptive skip modules for medical image segmentation can be searched more efficiently than the original IAC protocol suggests.
Instead of always running a full 200-epoch differentiable search, we track how the relative importance of competing operations on each edge of the IAC cell evolves during search and prune once those preferences stabilize.
The results show that the operations which survive discretization are already among the strongest candidates after only a small fraction of the search budget, and that the continuous architecture parameters stabilize long before the final epoch.

This behavior is compatible with LTH-style early winner emergence, but in the present paper its importance is methodological rather than purely theoretical.
Our JS-based stability test quantifies when the per-edge distribution over operations stops changing significantly.
Once an edge is stable, continuing to train all operations on that edge adds little information, and discarding weak operations has minimal impact on the eventual discrete cell.

In addition, we observe that the speed of convergence varies across edges and datasets, and that different random initializations of the IAC weights can lead to slightly different final genotypes, again echoing LTH findings on the sensitivity of tickets to initialization and training dynamics.
These effects are modulated by the backbone and dataset: larger backbones such as ResNet-50 converge slightly more slowly, and highly informative datasets such as BraTS favor earlier identification of winning operations.

\textbf{Practical implications.}
From a practical perspective, the main implication is that full-length differentiable NAS searches are often unnecessary when targeting plug-in modules such as IACs.
Once operation-level importance distributions stabilize and early winners are clearly separated, the search can be safely truncated.
Our JS-based criterion gives a simple, implementation-agnostic rule for deciding when this point has been reached.

The accelerated search reduces the computational barrier to using IACs in medical segmentation workflows.
Instead of running 200-epoch searches for each new dataset and backbone, practitioners can monitor JS divergence and prune low-importance operations as soon as stability is detected.
In our experiments this shortens the search from hundreds of epochs to a few dozen, and from many hours to at most a few hours of GPU time, without sacrificing segmentation accuracy.
Such savings are particularly relevant in hospitals and small labs where GPU time is scarce and rapid adaptation to new imaging protocols is critical.

Within the scope of this paper, these results also strengthen the case for skip-focused search in medical imaging: the gains are observed on public datasets, evaluated at the patient level, and retained in a standard 2-D nnU-Net pipeline, including experiments with and without augmentation.

\textbf{Limitations and future work.}
Our study has several limitations.
First, it focuses on a specific search space: a single IAC cell with spatially preserving operations inserted into 2-D U-Net skip connections.
Although we evaluate across four datasets and multiple backbones, the findings may not directly transfer to 3-D networks, non-U-shaped architectures or search spaces with more aggressive down-/up-sampling operations.

Second, the JS divergence threshold and pruning schedule are heuristic and hand-tuned.
While our choices work robustly across the studied datasets, more systematic methods for adaptive thresholding or Bayesian stopping rules could further improve reliability.
In particular, future work could explore fully training-free or low-cost proxies that predict stability based on early gradients or weight statistics, without requiring repeated full forward passes.

Third, the experimental design is primarily focused on segmentation quality and search cost.
We do not explicitly evaluate robustness to domain shift, calibration of prediction maps or energy consumption on edge devices, and the backbone-grid experiments use a single seed per backbone (multi-seed evaluation is restricted to the nnU-Net setting).
Since the IAC cell slightly increases inference cost, a more comprehensive study could combine our accelerated search with downstream efficiency optimizations such as quantization, structured pruning or hardware-aware constraints.

\textbf{Conclusion.}
We have shown that operation choices inside Implantable Adaptive Cells stabilize early enough to support shorter search for medical image segmentation.
By quantifying this stability with JS divergence and designing a simple on-the-fly pruning strategy (IAC-LTH), we reduce the IAC search cost by factors between roughly $3\times$ and $16\times$ in practice, while preserving or improving patient-level segmentation accuracy across four diverse public medical datasets and both plain U-Net and 2-D nnU-Net pipelines.
These findings support adaptive skip-module search as a more practical option for biomedical image segmentation when computational budget is limited.

% To print the credit authorship contribution details
\printcredits

%% Loading bibliography style file
%\bibliographystyle{model1-num-names}
\bibliographystyle{cas-model2-names}

% Loading bibliography database
\bibliography{sn-bibliography}

% Biography
%\bio{}
% Here goes the biography details.
%\endbio

%\bio{pic1}
% Here goes the biography details.
%\endbio

\end{document}